\documentclass[11pt]{article}

\usepackage{acl}

\usepackage{times}
\usepackage{latexsym}
\usepackage{dblfloatfix}
\usepackage{subcaption}
\usepackage{xcolor}
\usepackage{placeins}
\usepackage{xcolor}
\usepackage{hyperref}
\usepackage{placeins}
\usepackage{float}

\usepackage{fvextra}
\usepackage[T1]{fontenc}

\usepackage[utf8]{inputenc}

\usepackage{microtype}

\usepackage{inconsolata}

\usepackage{graphicx}
\usepackage{enumitem}
\usepackage{amsmath}
\usepackage{amssymb}
\definecolor{mygreen}{RGB}{16,110,16}
\definecolor{myyellow}{RGB}{214,214,20}
\definecolor{myred}{RGB}{228,26,28}
\definecolor{vocabcolor}{RGB}{128,0,128}
\usepackage[most]{tcolorbox}
\definecolor{basecolor}{RGB}{0,102,204} 
\definecolor{advcolor}{RGB}{200,0,0}           
\definecolor{domaincolor}{RGB}{0,153,76}       
\definecolor{compcolor}{RGB}{153,102,0}        
\definecolor{lengthcolor}{RGB}{128,0,128} 
\usepackage{booktabs}
\usepackage{multirow}
\title{Evaluating Multi-Dimensional Generalization of Large Language Models in Temporal Extraction Tasks}

\NewDocumentCommand\hficon{}{\includegraphics[scale=0.05]{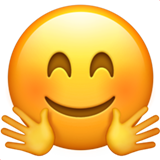}}
\NewDocumentCommand\githubicon{}{\includegraphics[scale=0.025]{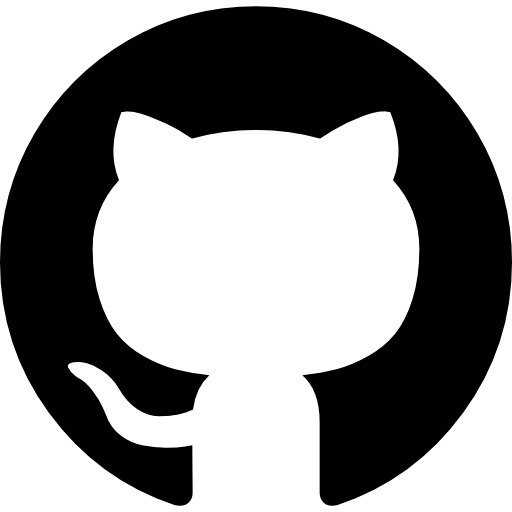}}

\author{
\textbf{Fahmid Shahriar Iqbal}$^{1}$,
\textbf{Ritam Dutt}$^{2}$,
\textbf{Soumitra Das}$^{3}$, \\
\textbf{Arnav Verma}$^{4}$,
\textbf{Sagnik Ray Choudhury}$^{1}$ \\[1mm]
$^{1}$University of North Texas,
$^{2}$Carnegie Mellon University \\
$^{3}$University of Massachusetts Amherst,
$^{4}$New York University \\[1mm]
\texttt{fahmidshahriar.iqbal@unt.edu},
\texttt{rdutt@andrew.cmu.edu},
\texttt{soumitradas@umass.edu} \\
\texttt{av4445@nyu.edu},
\texttt{sagnik.raychoudhury@unt.edu}
}

\newcommand{\conf}[1]{$\mathcal{M}_{#1}$}
\newcommand{\timex}{\textsc{timex}}
\newcommand{\eventx}{\textsc{eventx}}
\newcommand{\dv}[1]{\textcolor{vocabcolor}{#1}}

\newcommand\blfootnote[1]{%
  \begingroup
  \renewcommand\thefootnote{}%
  \footnote{#1}%
  \addtocounter{footnote}{-1}%
  \endgroup
}

\begin{document}
\maketitle
\begin{abstract}
Time and event expression extraction are fundamental temporal reasoning tasks, but the problem remains difficult due to annotation ambiguity, domain sensitivity, and unstable model behavior.
Existing evaluations focus on in-domain performance, offering limited insight into reliability under distribution shifts.
We evaluate multiple model configurations across families, architectures, and reasoning strategies over four dimensions of generalization, examining transfer from base performance, cross-dimensional correlations, and the effects of scale, architecture, and prompting. This provides a systematic study of how prompted LLMs generalize in time and event expression extraction tasks.
We find that strong base-task performance generally predicts better generalization. However, this relationship weakens under substantial distribution shifts. 
Inductive prompting performs most consistently across domain shift, adversarial perturbations, compositionality, and length increase, while gains from scale, architecture, and deductive and abductive prompting strategies are uneven and dimension-specific. 
We conclude that LLM generalization in temporal extraction tasks cannot be predicted from any single dimension alone and cannot be reliably inferred from in-domain or single-dimension evaluations, highlighting the need for reasoning strategies that generalize across dimensions.
\end{abstract}

\blfootnote{
\footnotesize
Code: \githubicon \href{https://github.com/nlpatunt/mdg-nlp.git}{GitHub}
\quad
Datasets: \hficon \href{https://huggingface.co/mdg-nlp/datasets}{HF}
\quad
Models:  \hficon \href{https://huggingface.co/mdg-nlp/models}{HF}
}

\section{Introduction}

Temporal reasoning in NLP aims to infer temporal structure from text, including events, time expressions, and the relations that order them over time \citep{pustejovsky2003timebank, ning2018multi}. Time and event expression extraction is a necessary first step in that process: in the sentence ``Monday the company announced a merger,'' a system must identify ``Monday'' and ``announced'' before determining when the announcement occurred. 

Low inter-annotator agreement \citep{cassidy2014annotation,styler2014temporal} in such span-level annotation points to the inherent difficulty of the problem. LLMs are increasingly applied to this problem, but recent work shows they are often inconsistent and unreliable \citep{qiu2024large,kougia2024analysing}. Consequently, robust extraction remains a critical bottleneck for reliable temporal reasoning, and in-domain evaluation alone is insufficient to assess reliability. Instead, models must be tested for generalizability.


Prior work argues for testing models in multidimensional generalization settings, covering different types of distributional shift and compositional variation \citep{hupkes2023taxonomy}. Such multi-dimensional generalization is crucial in temporal extraction, where deployed models must handle domain shifts, perturbations, and compositional complexity \citep{dutt2024investigating}.


\begin{figure}[ht]
    \centering
    \includegraphics[width=0.96\columnwidth]{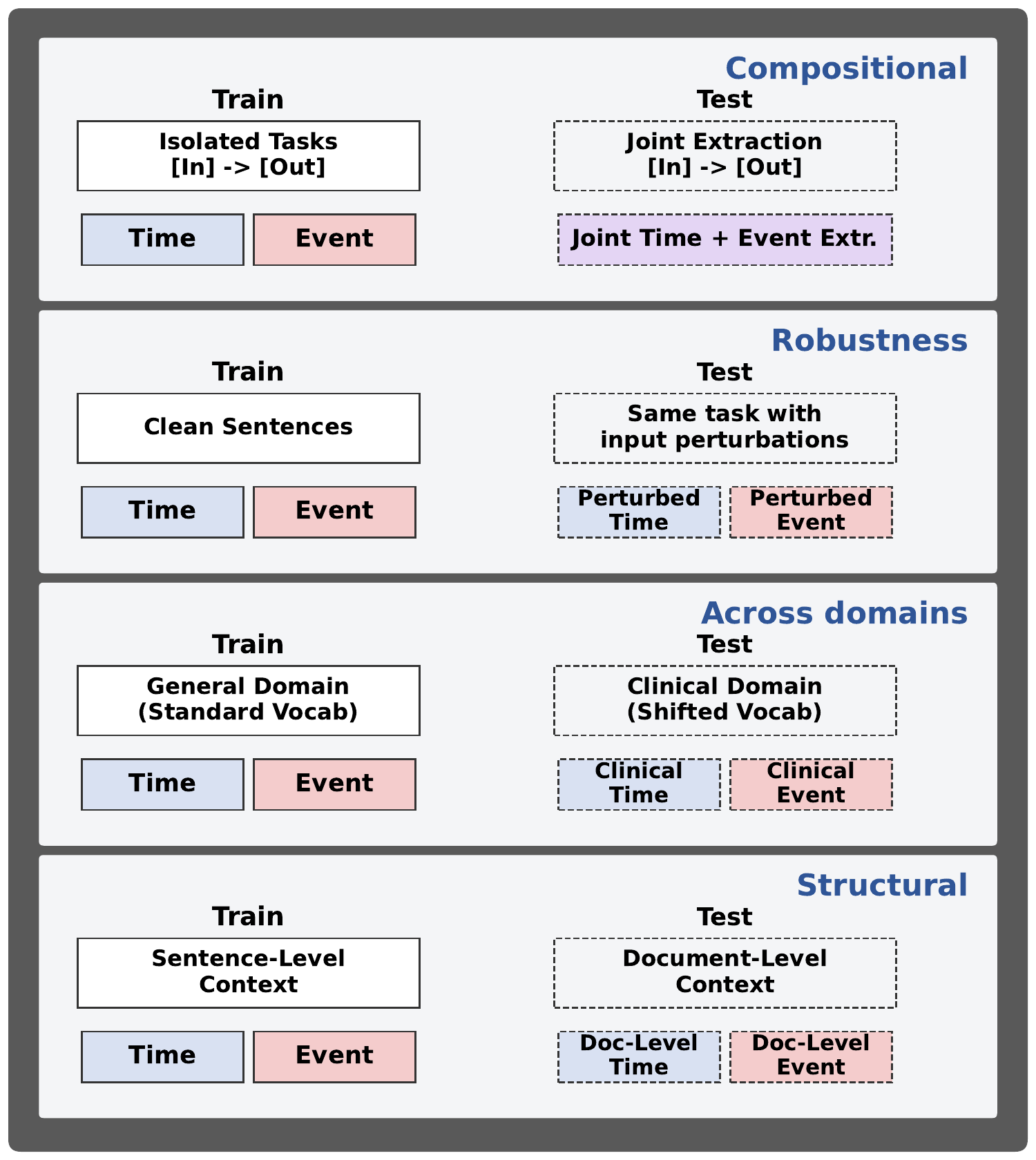}
    \caption{Generalization Dimensions.}
    \label{fig:dataset-generalization}
  
\end{figure}

\begin{figure}[ht]
    \centering
    \vspace{6pt}
    \includegraphics[width=\columnwidth]{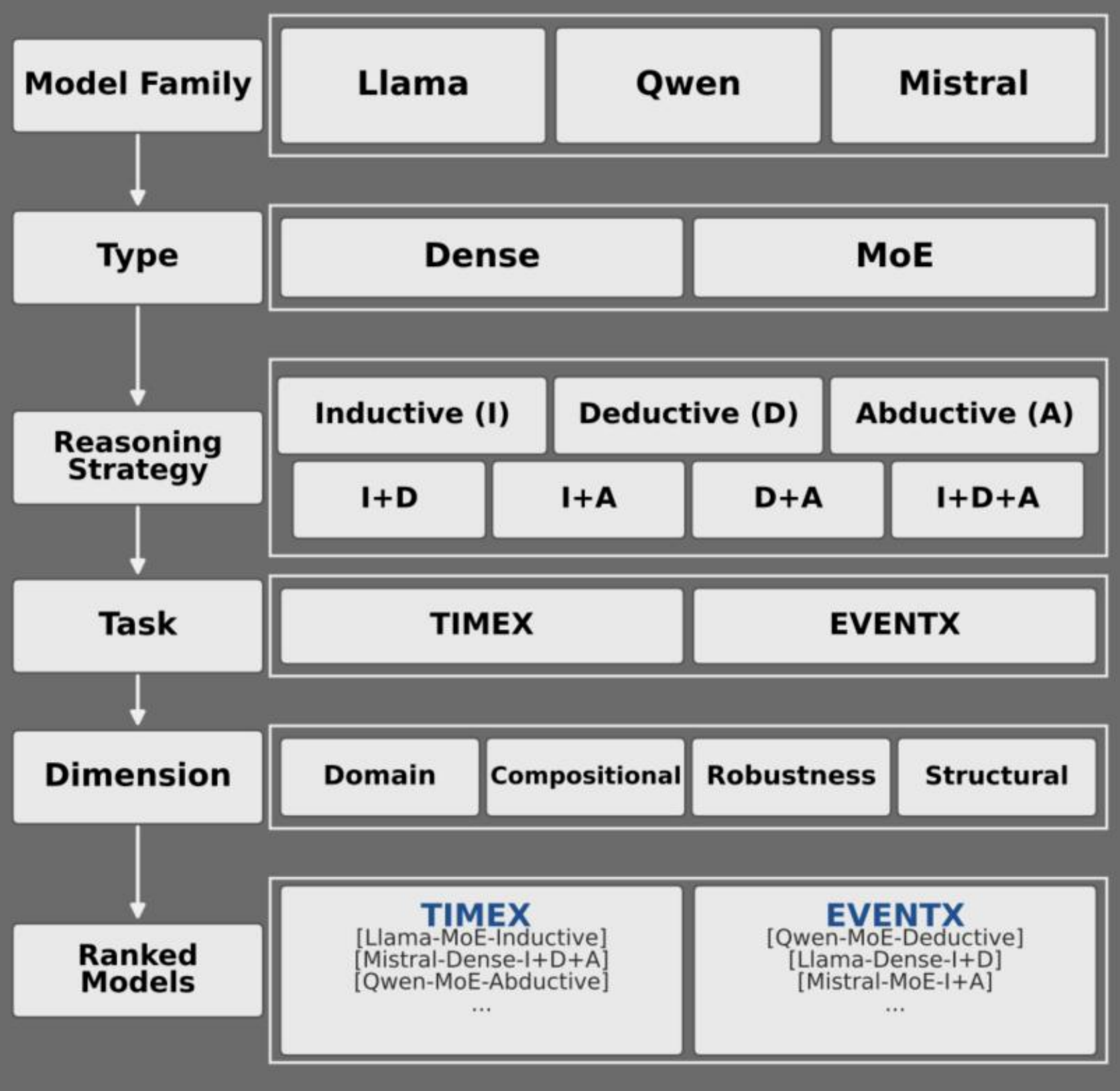}
    \caption{Pipeline for evaluation across tasks, architectures, and reasoning strategies.}
    \label{fig:experimental-setup}
    
\end{figure}

We present a systematic study of the multi-dimensional generalization of recent LLMs in time and event expression extraction tasks (\timex\ and \eventx). Following prior work \citep{hupkes2023taxonomy, dutt2024investigating}, we consider four dimensions of generalization (Figure \ref{fig:dataset-generalization}). \textbf{Domain generalization} measures models' ability to handle distribution shifts across domains with different linguistic characteristics (e.g., general vs. clinical text). \textbf{Robustness generalization} tests models' robustness to meaning-preserving perturbations. For \textbf{Compositional generalization}, the task is extended from extracting time and event expressions independently to extracting them jointly. Finally, for \textbf{Structural generalization}, the input context varies from sentence-level to document-level. These dimensions form one axis of variation, capturing how model behavior changes under different types of distribution shifts.

In parallel, we consider reasoning strategies as another axis of variation in model behavior. Reasoning strategies define how inputs are framed and structured under different prompting conditions, including deductive, inductive, and abductive reasoning, as well as combinations of these \citep{liu2024incomplete}, and performance under one strategy does not necessarily transfer to others. Together, multiple generalization dimensions and reasoning strategies define two complementary axes that enable a systematic evaluation of model configurations.
 
We evaluate models from three families: LLaMA \citep{touvron2023llama}, Qwen \citep{bai2023qwen}, and Mistral \citep{jiang2024mixtral}, and for each, we use $2$ architectural variants: Dense and Mixture-of-Experts (MoE). This, combined with the $7$ reasoning strategies, yields a total of 42 model configurations (\conf{i}s). Each \conf{i} is evaluated on both temporal and event extraction tasks across all generalization dimensions. With this extensive setup (Figure \ref{fig:experimental-setup}), we ask the following research questions:
\begin{itemize}[leftmargin=*, itemsep=0pt]
    \item \textbf{RQ1: } If a model configuration \conf{i} performs well on base extraction tasks, does the performance consistently extend across all generalization dimensions? If not, practitioners risk selecting \conf{i}s that perform well in standard evaluations but fail in practice (\S\ref{sec:rq1}).
    \item \textbf{RQ2: } To what extent are different generalization dimensions correlated? Specifically, if a configuration performs well in one dimension, how likely is it to perform well in others? Unlike RQ1, which studies transfer from base tasks, RQ2 examines interdependence among generalization dimensions (\S\ref{sec:rq2}).
    \item \textbf{RQ3: } How do prompting and model factors affect generalization across dimensions? We examine the effects of reasoning strategy, model architecture (Dense vs. MoE), and model size. These factors are orthogonal to the generalization dimensions studied in RQ1 and RQ2, and their effects may vary across dimensions (\S\ref{sec:rq3}).
\end{itemize}

We show that strong base-task performance generally predicts multi-dimensional generalization, except under substantial distribution shifts. Among reasoning strategies, inductive ones perform well on base tasks as expected, but they also exhibit strong multi-dimensional generalization and greater cross-dimensional consistency. Surprisingly, combining instructions (deductive) with demonstrations (inductive) does not meaningfully improve generalization, whereas prompting the model to produce explanations (abductive) does. MoE models outperform their dense counterparts on base tasks and across many dimensions of generalization, but this advantage diminishes under substantial distributional shifts. Finally, scaling model size improves adversarial and structural generalization but yields no significant gains for domain or compositional ones. Compared to prompted LLMs, fine-tuned PLMs used as baselines exhibit more consistent cross-dimensional behavior, though this consistency does not always reflect strong performance. These findings call for reasoning strategies and evaluation frameworks that explicitly target robust, cross-dimensional generalization rather than assuming it follows from strong base-task performance.

\section{Experimental Setup}

\subsection{Tasks \& Datasets}
Temporal reasoning in natural language requires identifying what happened and when. This involves extracting event expressions (\eventx), which capture semantic occurrences (e.g., actions, states, or reporting events), and temporal expressions (\timex), which capture dates, times, durations, or relative references like today or last week. These span extraction problems are prerequisites for advanced temporal reasoning tasks such as ordering events, determining causality, or building timelines. Therefore, we focus on them.

\textbf{Base Corpus: TimeBank.} We derive the base datasets for both \timex\ and \eventx\ tasks from the TimeBank 1.2 corpus \citep{pustejovsky2003timebank}. The corpus is converted into sentence-level instances by segmenting documents and retaining only sentences that contain at least one target expression, with annotations projected accordingly. We remove all additional annotation attributes (e.g., normalization for temporal expressions and linguistic features for events). Each dataset consists of sentence--target pairs, with a raw sentence as input and a set of textual spans as output.

\textbf{Adversarial Datasets: \textit{Adv} \& \textit{Adv\_P}.} We construct adversarial variants of the base datasets using a modified SeqAttack framework \citep{simoncini2021seqattack}, adopting two widely used attack strategies: TextFooler \citep{jin2020bert} and DeepWordBug \citep{gao2018black}. TextFooler is a word-level attack that iteratively replaces important tokens with semantically similar synonyms while preserving the overall meaning. In contrast, DeepWordBug operates at the character level, introducing small perturbations (e.g., swaps, insertions, deletions) that disrupt tokenization and degrade model performance. For \timex\, we use TextFooler and DeepWordBug to generate adversarial examples for each sentence and merge the perturbations. For \eventx, we restrict the perturbation to TextFooler, as event triggers are sensitive to local syntactic context. These datasets are referred to as \textit{Adv}.

We also construct a prompt-based adversarial variant (\textit{Adv\_P}) using GPT-4.1\citep{xu2023llm}. We prompt the model to introduce lexical substitutions, character-level noise, and structural modifications, while preserving sentence meaning. We validate that all target expressions remain unchanged.

\textbf{Domain-Shift: Clinical Dataset (\textit{T2C}) and Vocabulary Replacement Dataset (\textit{Voc}).} To evaluate cross-domain generalization, we construct datasets (\textit{T2C}) using the 2012 i2b2 Clinical Temporal Relations corpus \citep{sun2013annotating}. Compared to TimeBank news data, clinical narratives contain specialized medical vocabulary, abbreviations, and telegraphic sentences. To ensure consistency, we apply the same preprocessing pipeline as in the base datasets. 

We create another domain dataset (\textit{Voc}) by systematically replacing selected tokens (e.g. entity-like tokens, rare words, and domain-specific terms) with synthetic or semantically related alternatives, preserving sentence structure and label  \citep{ribeiro2020beyond}. Compared to \textit{T2C}, \textit{Voc} represents a more controlled and localized shift in vocabulary rather than a full domain change.

\textbf{Structural Generalization: Document Level Dataset (\textit{Len}).} Structural generalization refers to models' ability to effectively handle inputs of varying or increased length. We revert to the original documents of the TimeBank corpus for this purpose.

\textbf{Compositional Dataset (\textit{Comp}).} Compositional generalization captures a model's ability to solve more complex tasks by systematically combining solutions to smaller, simpler sub-tasks it has learned. To mimic this, we repurpose the TimeBank dataset so that the models must jointly extract time and event expressions.

See \S \ref{sec:app-datasets} for a detailed discussion, and examples from all base and generalization datasets (Figures \ref{fig:timex_examples} and \ref{fig:event_examples}).

\subsection{Models}
Based on prior work highlighting trade-offs between throughput, latency, and memory efficiency during inference \citep{agrawal2024taming,fu2024serverlessllm}, we organize models into two computational tiers.

\textbf{Medium Tier (Dense Architectures, 16--24 GB VRAM)}
consists of compact dense models that run efficiently on a single GPU. We include \href{https://openrouter.ai/mistralai/ministral-8b-2512}{Mistral 8B}, \href{https://openrouter.ai/meta-llama/llama-3-8b-instruct}{Llama 3 8B Instruct}, and \href{https://openrouter.ai/qwen/qwen-2.5-7b-instruct}{Qwen2.5 7B Instruct}, representing small-scale dense architectures with comparable parameter sizes.

\textbf{Heavy Tier (Mixture-of-Experts Models, $\geq$ 48 GB VRAM)}
includes large-scale mixture-of-experts (MoE) models with conditional routing, enabling us to study the effect of expert activation on generalization. We evaluate \href{https://openrouter.ai/meta-llama/llama-4-maverick}{Llama 4 Maverick} (17B active parameters), \href{https://openrouter.ai/qwen/qwen3-235b-a22b-2507}{Qwen3 235B A22B} (22B active parameters), and \href{https://openrouter.ai/mistralai/mixtral-8x7b-instruct}{Mixtral 8x7B} Instruct (13B active parameters).

\textbf{Scaling.} We conduct a controlled study on the Mistral dense model family: \href{https://openrouter.ai/mistralai/ministral-3b-2512}{Mistral 3B}, \href{https://openrouter.ai/mistralai/mistral-small-3.2-24b-instruct}{Mistral Small 3.2 (24B)}, and \href{https://openrouter.ai/mistralai/mistral-large}{Mistral Large (123B)}, to examine how increasing parameter count impacts generalization.

\subsection{Reasoning strategy}

Reasoning strategy \citep{liu2024incomplete} forms one of the primary axes of our study, alongside the different generalization dimensions. We evaluate each model under seven strategies: inductive ($\mathcal{I}$), deductive ($\mathcal{D}$), abductive ($\mathcal{A}$), and their combinations, across both \timex\ and \eventx\ tasks.

\textbf{Inductive} reasoning corresponds to in-context learning, in which the task is implicitly specified by examples from which the model infers patterns \citep{brown2020language}. We implement this via few-shot prompting (\S~\ref{sec:Inductive}), where models infer extraction patterns from annotated examples. For consistency, examples are always drawn from the base dataset, and the best-performing number of exemplars ($k \in {3,5,8}$) is fixed across all settings. In contrast, the \textbf{deductive} reasoning does not use examples; instead, the task is explicitly specified as a set of rules or a natural language instruction, and the model applies these rules to produce outputs. Specifically, we provide a clear definition of valid time and event expressions (\S~\ref{sec:Deductive}). \textbf{Abductive} reasoning involves generating intermediate hypotheses or explanations to guide prediction \citep{wei2022chain}. In our setup, we implement this by generating candidate spans with reasoning steps and selecting those that best justify the extraction (\S~\ref{sec:Abductive}). We also consider combinations of reasoning strategies that integrate multiple paradigms within a single prompt. For instance, $\mathcal{D+I}$ incorporates both instructions and few-shot examples, enabling the model to align explicit rules with observed patterns. 
These hybrid setups capture interactions between reasoning mechanisms that are not observable in isolated settings. 

\subsection{Evaluation Metrics}

We evaluate the quality of predicted expressions using a Hungarian word-level matching metric and report the micro F1 score as our primary metric. 


Let $\mathcal{P} = \{p_1, \dots, p_m\}$ denote the set of predicted expressions and $\mathcal{G} = \{g_1, \dots, g_n\}$ denote the set of gold expressions for a given instance. For each pair $(p_i, g_j)$, we compute a similarity score based on the word-level Longest Common Subsequence (LCS), normalized by span length: $\text{sim}(p_i, g_j) = \frac{\text{LCS}(p_i, g_j)}{\max(|p_i|, |g_j|)}$ where $|p_i|$ and $|g_j|$ denote the number of tokens in the spans.

We construct a similarity matrix $\mathbf{S} \in \mathbb{R}^{m \times n}$ with entries $S_{ij} = \text{sim}(p_i, g_j)$ and compute an optimal one-to-one matching $\mathcal{M} \subseteq \mathcal{P} \times \mathcal{G}$ that maximizes the total similarity: $\mathcal{M} = \operatorname*{arg\,max}_{\text{matching}} \sum_{(i,j) \in \mathcal{M}} S_{ij}$ using the Hungarian algorithm \citep{kuhn1955hungarian}. Precision $\frac{\sum_{(i,j) \in \mathcal{M}} S_{ij}}{|\mathcal{P}|}$ and Recall $\frac{\sum_{(i,j) \in \mathcal{M}} S_{ij}}{|\mathcal{G}|}$ are then defined based on the matched overlaps.

This approach prevents multiple matches to the same expression, providing a more rigorous evaluation similar to span-based protocols used in the CoNLL-2003 Shared Task \citep{sang2003introduction} and SemEval-2013 Task 9.1 \citep{segura2013semeval}.

\section{Results}
\label{sec:results}
All prompted LLMs are accessed via the OpenRouter API\footnote{\url{https://openrouter.ai/}} with default settings and temperature 0, ensuring a consistent and reproducible inference setup. The detailed results are provided in \S \ref{sec:app-detailed-results}.

\subsection{RQ1: Transferability of Base Task Performance Across Dimensions}
\label{sec:rq1}

\begin{figure*}[t!]
    \centering
    \includegraphics[width=0.9\textwidth]{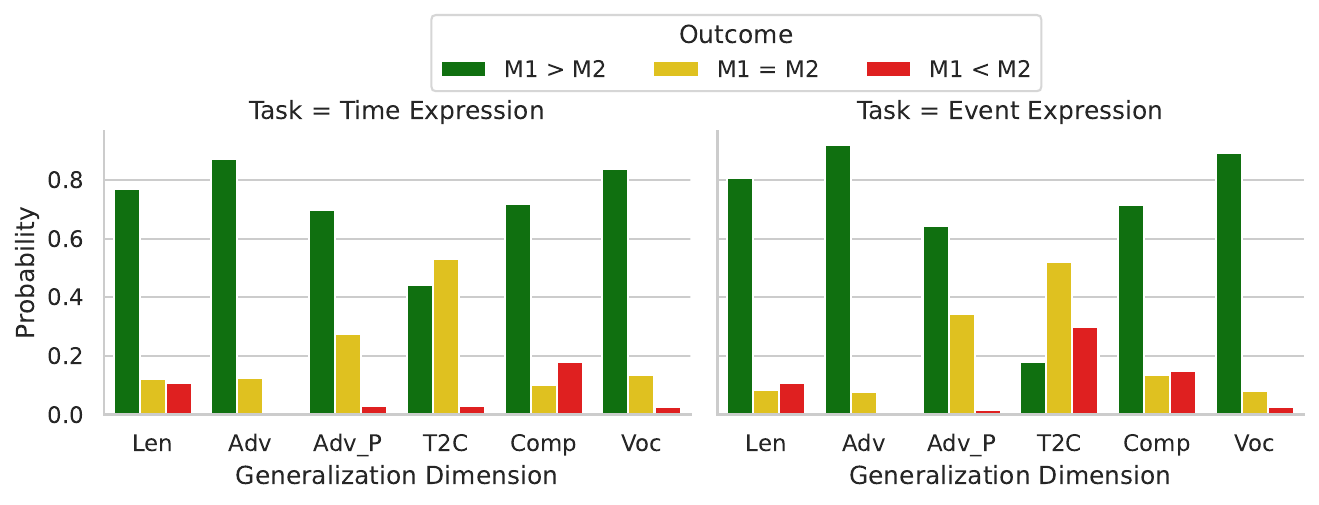}
    \caption{If \conf{1} is significantly better than \conf{2} on the base case, what is the probability that the advantage will hold for different generalization dimensions?}
    \label{fig:sig-diff}
    \vspace{-8pt}
\end{figure*}

\begin{figure*}[t!]
    \centering
    \includegraphics[width=0.9\textwidth]{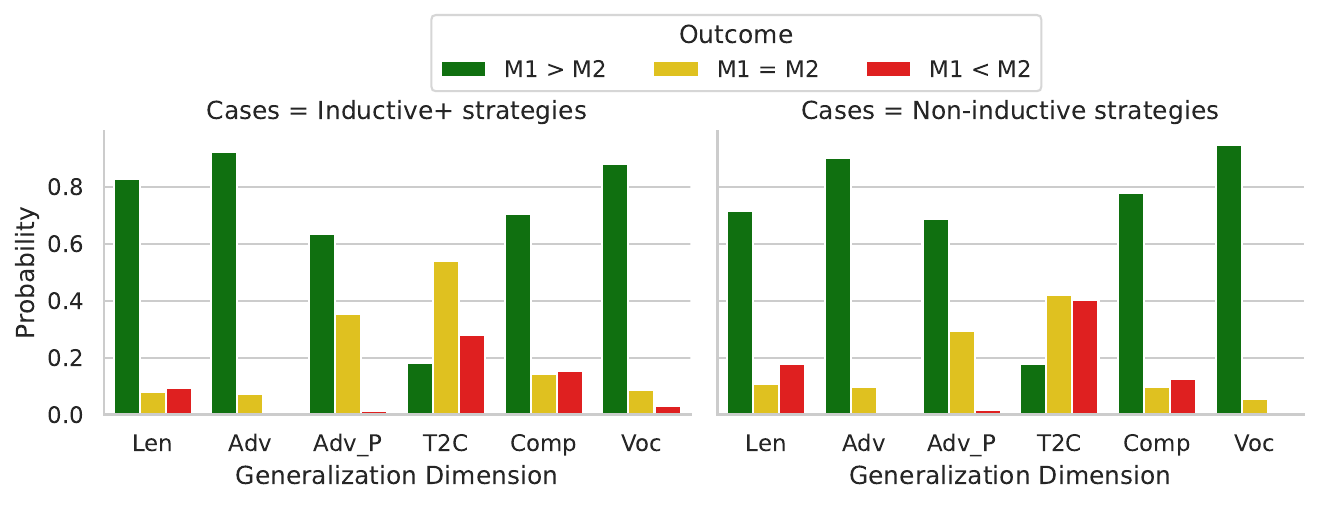}
    \caption{Generalization probabilities for \eventx, split into inductive+ (left) and non-inductive (right) strategies.}
    \label{fig:ind-vs-nonind-sr}
    \vspace{-8pt}
\end{figure*}

We evaluate whether strong performance on base extraction tasks translates across all generalization dimensions. To do so, we examine whether pairwise orderings between configurations, established on the base dataset, persist across generalization dimensions. Specifically, consider two configurations \conf{i} and \conf{j} where $i$ is significantly better than $j$ on the \textit{base} dataset. What is the probability that \conf{i} will be significantly better, worse, or almost equal to \conf{j} in \textit{generalization} datasets? Figure \ref{fig:sig-diff} shows the results for both \timex\ and \eventx\ datasets. 
Statistical significance ($\text{p} < 0.05$) is determined using a paired bootstrap test \citep{berg2012empirical}.

First, no configuration achieves a $100\%$ probability. Based on base-task performance, no single \conf{i} consistently guarantees better generalization than the others. However, for \timex, if an \conf{i} performs better than \conf{j} on the base task, it is highly likely to generalize better than \conf{j} on adversarial, length, and compositional datasets. If the domain generalization dataset has a very strong distribution shift (e.g., \textit{T2C}), then the models will have a much lower probability of generalizing; otherwise, e.g., with a simple vocabulary shift (\textit{Voc}), the generalization trend will hold. \eventx\ results are also similar, but the \textit{T2C} generalization probability reduces even further.

Can the \textit{T2C} anomaly be attributed to \textit{inductive+} ($\mathcal{I}$, $\mathcal{A+I}$, $\mathcal{D+I}$, $\mathcal{A+D+I}$) reasoning strategies? These strategies generally yield better performance on the base dataset (see Figure \ref{fig:app-ind-plus-vs-non-inductive}, \S \ref{sec:app-ind-plus-vs-non-inductive}), likely because they use exemplars drawn from the same distribution. However, this use of in-domain examples may negatively affect generalization. In Figure \ref{fig:ind-vs-nonind-sr} we plot these probabilities separately for inductive+ and non-inductive strategies for \eventx\ datasets. Interestingly, the use of in-domain examples boosts generalization, even for difficult domain shifts such as \textit{T2C}.

\subsection{RQ2: Correlation Across Generalization Dimensions}
\label{sec:rq2}

\begin{figure*}[t!]
    \centering

    \begin{subfigure}[t]{0.49\textwidth}
        \centering
        \includegraphics[width=\linewidth]{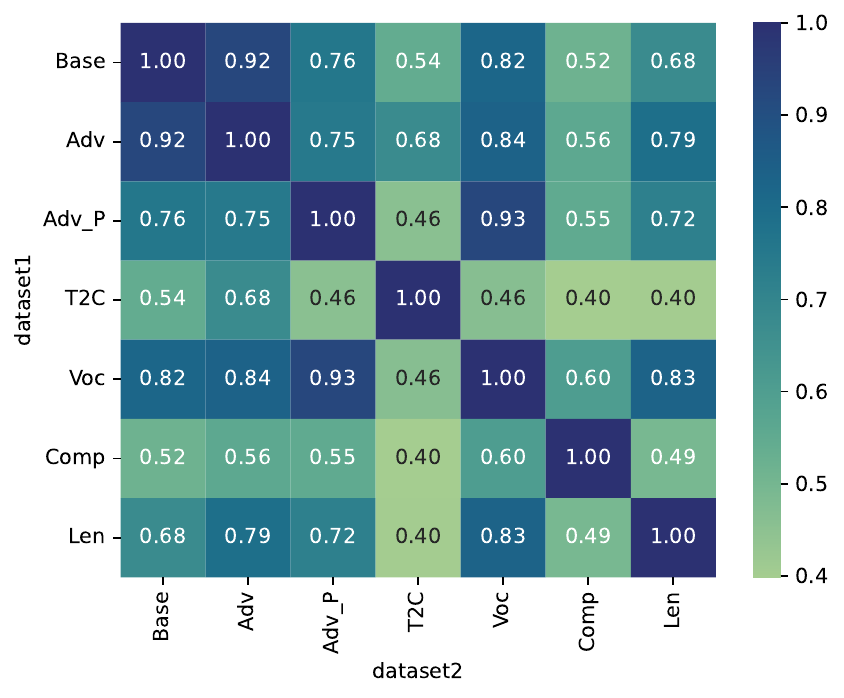}
        \caption{Time expression datasets}
        \label{fig:temp-extr-instwise-main}
    \end{subfigure}
    \hfill
    \begin{subfigure}[t]{0.49\textwidth}
        \centering
        \includegraphics[width=\linewidth]{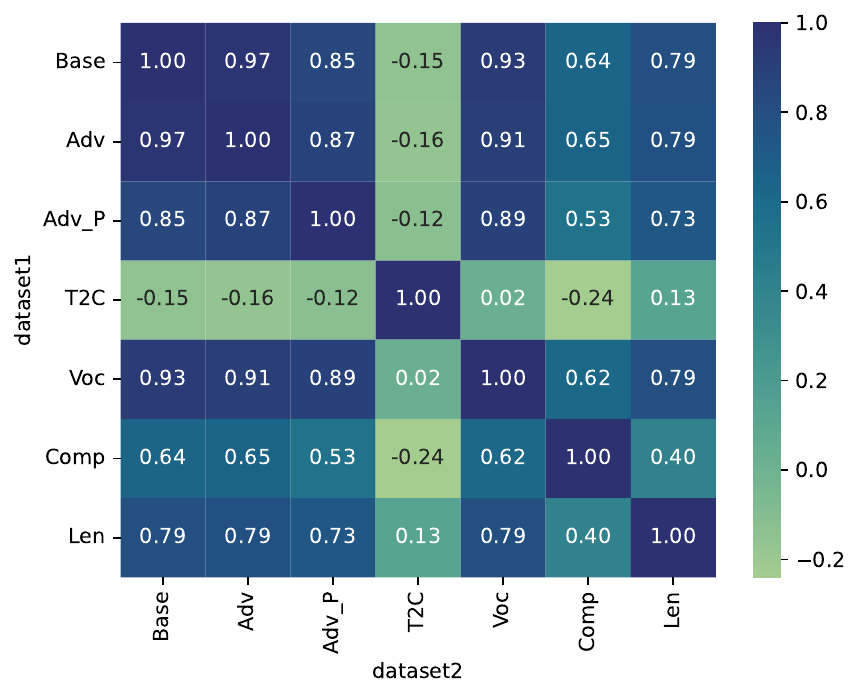}
        \caption{Event expression datasets}
        \label{fig:event-extr-instwise-main}
    \end{subfigure}
    \caption{Spearman's Rank Correlation $\rho$ between the source and the target datasets for different generalization dimensions: temporal and event expression extraction.}
    \label{fig:instwise-sr-corr}
    \vspace{-8pt}
\end{figure*}

For RQ2, we assess whether configurations (\conf{i}) maintain consistent relative performance across different generalization dimensions. We first rank all configurations separately on each dataset, then compute Spearman's rank correlation ($\rho$) between every pair of datasets to measure how similar these rankings are.
Figures \ref{fig:temp-extr-instwise-main} and \ref{fig:event-extr-instwise-main} show the results for \timex\ and \eventx\ tasks, respectively. 

For \timex\ datasets, model rankings show moderate-to-strong consistency across most generalization dimensions. Adversarial perturbation datasets are well correlated with each other as expected.\footnote{$\rho = 0.75$, and $\rho \geq 0.6$ indicates strong to very-strong correlation \cite{statstutor_spearman}.} The domain-shift datasets (\textit{T2C} and \textit{Voc}) have lower but moderate correlations. Interestingly, adversarial generalization (across both datasets) is well correlated with domain and structural generalization, but not with compositional generalization. These trends tend to hold for event expression datasets, except for \textit{T2C}, which is, in fact, negatively correlated with other dimensions. A negative correlation implies that \conf{i}s that perform well on \textit{T2C} tend to perform worse on other dimensions, i.e., optimizing for this domain shift may harm generalization elsewhere. As before, we investigate whether this is due to sample selection in the inductive strategies. Figures \ref{fig:eventx-inductive-sr} and \ref{fig:eventx-noninductive-sr} (\S \ref{sec:app-corr-gen-dim}) show that this is not the case. The correlations, although low, are much higher in the inductive+ strategies than the non-inductive ones.

\begin{figure*}[t]
    \centering
    \includegraphics[width=\textwidth]{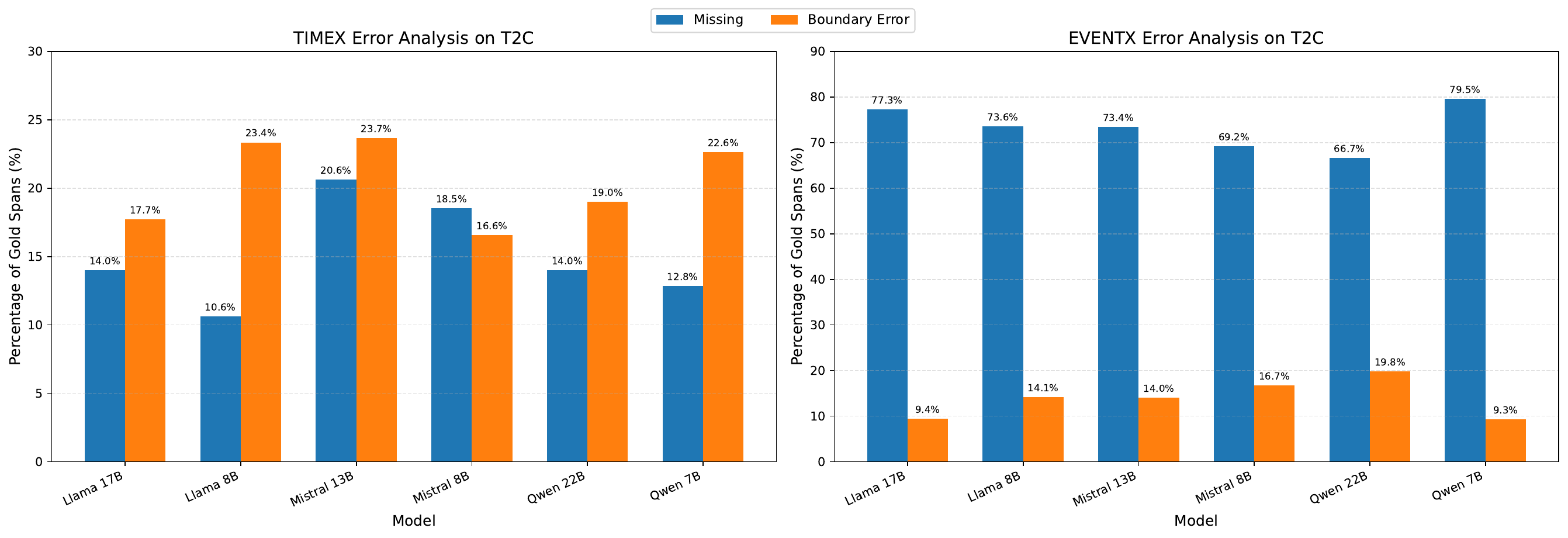}

    \captionsetup{width=\textwidth}
    \caption{Extraction error analysis on \textit{T2C}, pooled across all reasoning strategies.}
    \label{fig:t2c_error_analysis}
\end{figure*}

\begin{figure*}[t]
\centering

\begin{subfigure}[t]{0.49\textwidth}
    \centering
    \includegraphics[width=\linewidth]{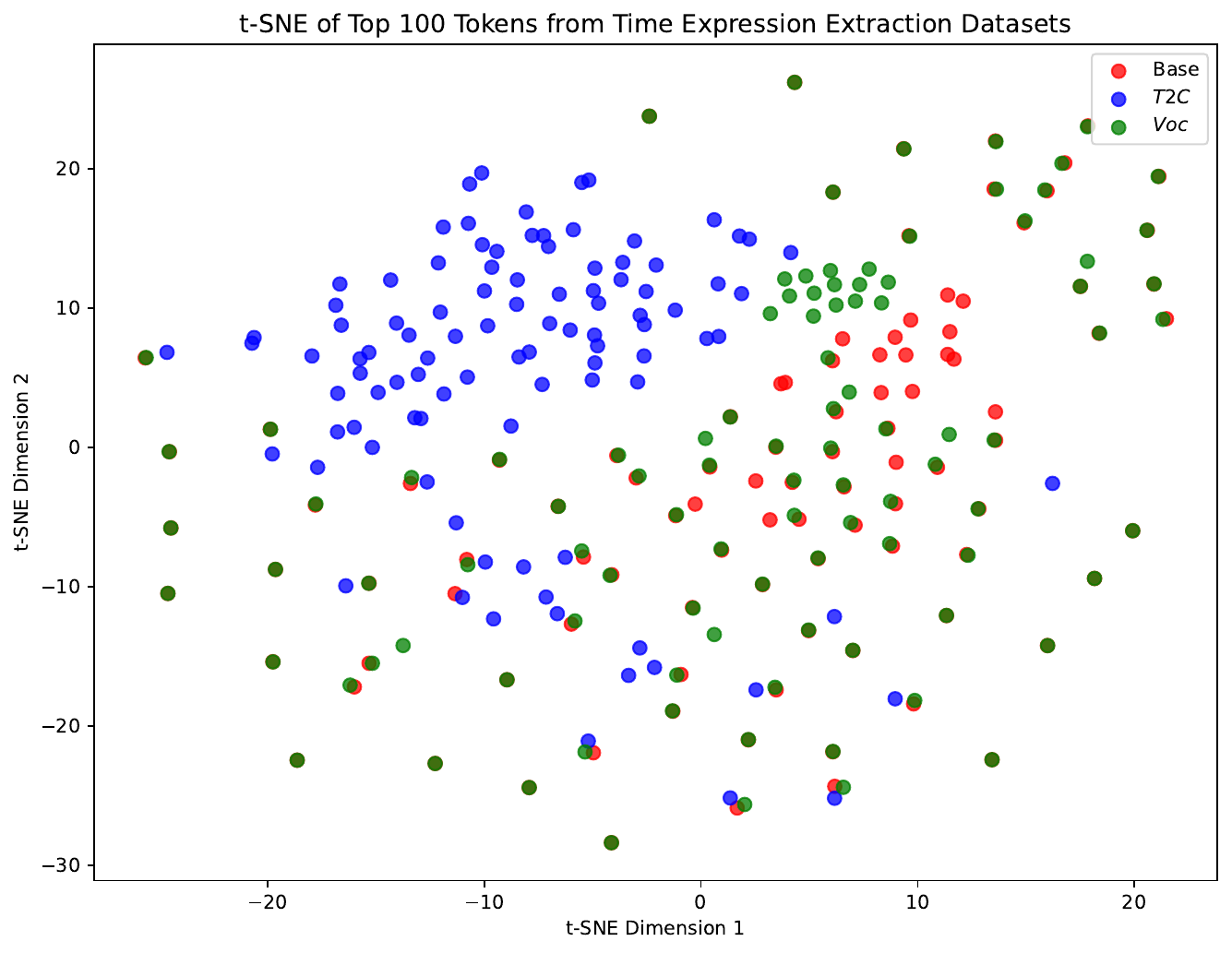}
    \caption{\timex\ datasets.}
    \label{fig:tsne_timex}
\end{subfigure}
\hfill
\begin{subfigure}[t]{0.49\textwidth}
    \centering
    \includegraphics[width=\linewidth]{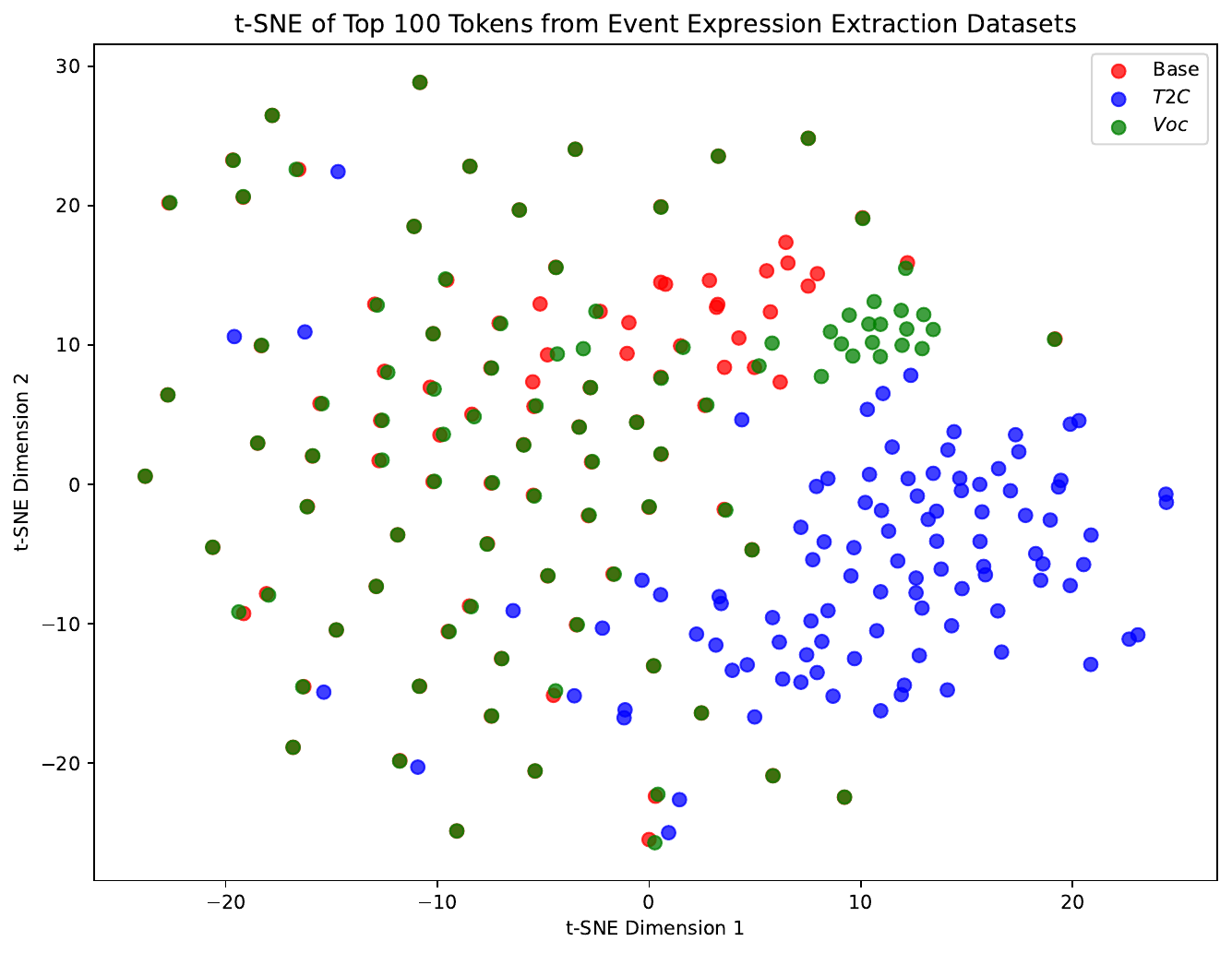}
    \caption{\eventx\ datasets.}
    \label{fig:tsne_eventx}
\end{subfigure}

\caption{
t-SNE projections of final-layer \texttt{roberta-base} token embeddings from the Base, \textit{T2C}, and \textit{Voc} datasets.
}
\label{fig:tsne_combined}
\end{figure*}

To further characterize \textit{T2C} 's distinct behavior, we analyze its extraction errors in Figure~\ref{fig:t2c_error_analysis}. For \timex, errors consist of both missing and boundary-mismatched spans, whereas \eventx\ errors are dominated by missing spans, with 66.7--79.5\% of gold event expressions missed across models.
The t-SNE projections in Figure~\ref{fig:tsne_combined}, based on final-layer \texttt{roberta-base} embeddings of the 100 most frequent tokens, further illustrate the distributional differences among Base, \textit{T2C}, and \textit{Voc}. Greater separation is observed between \textit{T2C} and Base tokens than between \textit{Voc} and Base, suggesting a stronger lexical and contextual shift. We hypothesize this to be a reason for \textit{T2C} behaving differently from other generalization dimensions. \\
\textbf{In summary}, a strong base-task performance of \conf{i} is generally a good indicator of multi-dimensional generalization, with some exceptions such as significant distribution shifts (e.g., \textit{T2C}). Most importantly, inductive+ strategies, where few-shot examples are sampled from the base datasets, not only boost \textit{base performance} (Figure \ref{fig:app-ind-plus-vs-non-inductive}, \S \ref{sec:app-ind-plus-vs-non-inductive}) but also improve generalization. Furthermore, they maintain significantly higher cross-dimensional correlation than non-inductive approaches.

\subsection{RQ3: Effects of Reasoning Strategies, Model Architecture, and Scale}
\label{sec:rq3}

\begin{figure*}[t!]
    \centering
    \includegraphics[width=\textwidth]{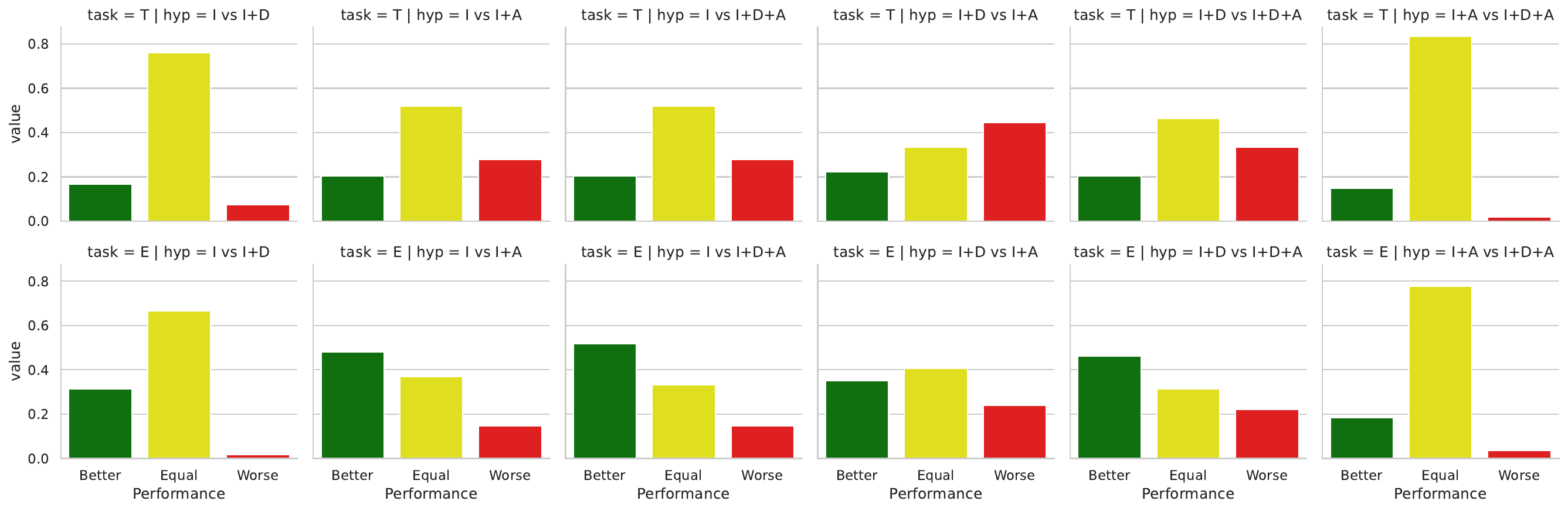}

    \caption{Percentage of models that perform significantly \textcolor{mygreen}{better}, \textcolor{myred}{worse}, or \textcolor{myyellow}{almost equal} for reasoning strategy $i$ vs.\ $j$ (\texttt{Hyp = i vs j}), summed over all generalization dimensions. The upper and lower panels show the results for \timex\ (\texttt{T}) and \eventx\ (\texttt{E}) tasks, respectively.}

    \label{fig:reasoning-strategy-time-event}
    \vspace{-8pt}
\end{figure*}

\textbf{Reasoning strategy.} The previous results indicate that inductive+ strategies are better for generalization. Following up, we investigate whether, among these strategies, a particular one, say, $\mathcal{D+I}$, is better than the others. Specifically, for any reasoning strategy pair ($r_i, r_j$) and a generalization dimension, say, \textit{T2C}, we note the percentage of models that perform significantly ($\text{p} < 0.05$) better, worse, or almost equal ($\text{p} \geq 0.05$) for $r_i$ than $r_j$. Figure \ref{fig:reasoning-strategy-time-event} shows the results when the results are averaged over \textit{all dimensions}. Results for individual dimensions are shown in Figures \ref{fig:reasoning-strategy-time-all-dim} and \ref{fig:reasoning-strategy-event-all-dim}, \S \ref{sec:app-reasoning-strategy}, and have similar trends.

For \timex, addition of deductive prompts, i.e., instructions, hampers generalization, but the addition of explanations improves it ($\mathcal{I}$ is better than $\mathcal{I+D}$, and $\mathcal{I+A}$ is better than $\mathcal{I+D}$). Perhaps most interestingly, $\mathcal{I+D+A}$, i.e., ``few-shot chain-of-thought'' is not the best reasoning strategy for generalization. For \eventx, the base inductive strategy ($\mathcal{I}$) is better than all other alternatives.

\begin{figure*}[t!]
    \centering
    \includegraphics[width=\textwidth]{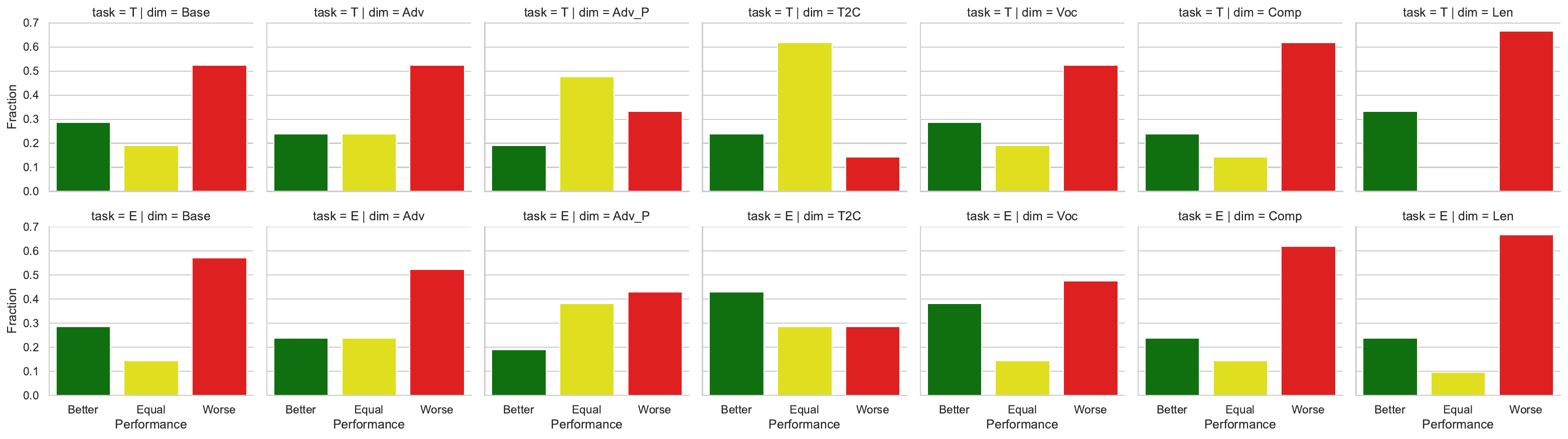}

    \caption{Percentage of Dense models that perform significantly \textcolor{mygreen}{better}, \textcolor{myred}{worse}, or \textcolor{myyellow}{almost equal} for the corresponding MoE model in a particular dimension (\texttt{dim = x}), summed over all reasoning strategies.}

    \label{fig:dense-v-moe}
    \vspace{-8pt}
\end{figure*}

\textbf{Model architecture.} For each generalization dimension, we report whether Dense models perform significantly better, worse, or nearly equal to the closest available MoE model from the same family. Exact official size-matched MoE counterparts are not available, so we use the closest comparable variants: Llama 4 Maverick, Mixtral, and Qwen MoE. Although these MoE models have larger active parameter counts, they remain broadly comparable in scale (Llama: 8B vs. 17B, Mistral: 8B vs. 13B, Qwen: 7B vs. 22B). Thus, worse MoE generalization \textit{despite larger active size} suggests limitations beyond parameter count. Figure \ref{fig:dense-v-moe} shows the results.

MoE models outperform Dense counterparts on base tasks, but their advantage weakens under larger shifts such as \textit{T2C} and \textit{Adv\_P}, where they perform similarly to Dense models, and on \eventx\ tasks, where they perform worse. This suggests that higher active parameter counts do not reliably translate into better domain generalization.

\begin{figure}[t]
    \centering
    \includegraphics[width=\columnwidth]{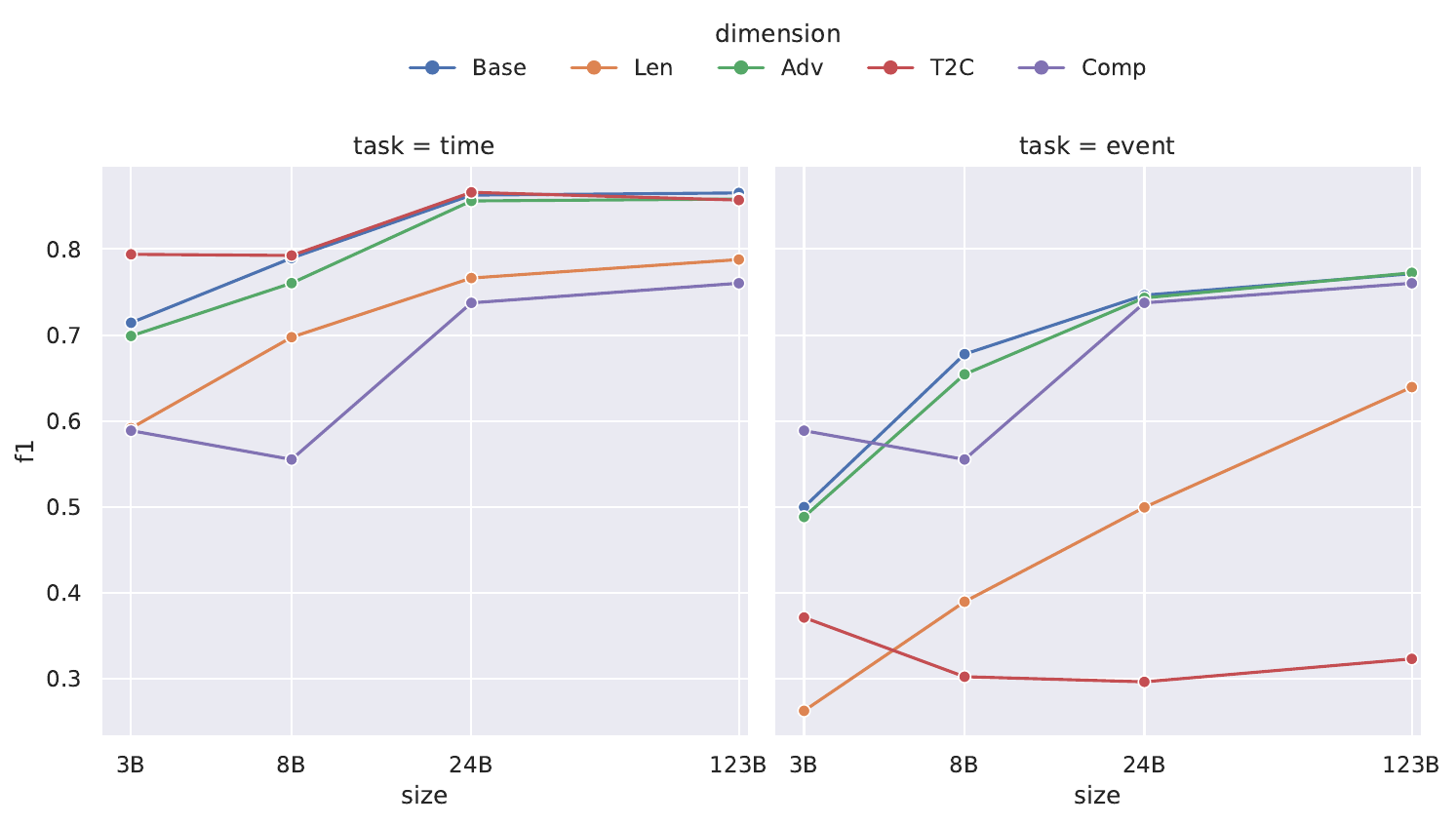}

    \caption{Generalization performance of Mistral Dense models with varying parameter sizes.}

    \label{fig:model-size}
    \vspace{-8pt}
\end{figure}

\textbf{Model size.} We investigate the effect of model size on generalization with 4 Dense Mistral models of sizes 3B to 123B and a subset of dimensions: a) \textit{Adv}, b) Length, c) \textit{T2C}, and d) Compositional. The results are presented in Figure \ref{fig:model-size}.

Model size correlates well with base-task performance and generalization, but the gains are not uniform across all dimensions. Performance improves monotonically in base, and the same behavior is observed in \textit{Adv} and Length. However, for \textit{T2C} and compositional, the performance difference between smaller changes (3B $\rightarrow$ 8B) is non-significant. Surprisingly, for \eventx, model size has no effect on domain generalization. In summary, scaling alone does not guarantee consistent gains across all generalization dimensions.

\begin{figure}[t]
    \centering
    \includegraphics[width=\columnwidth]{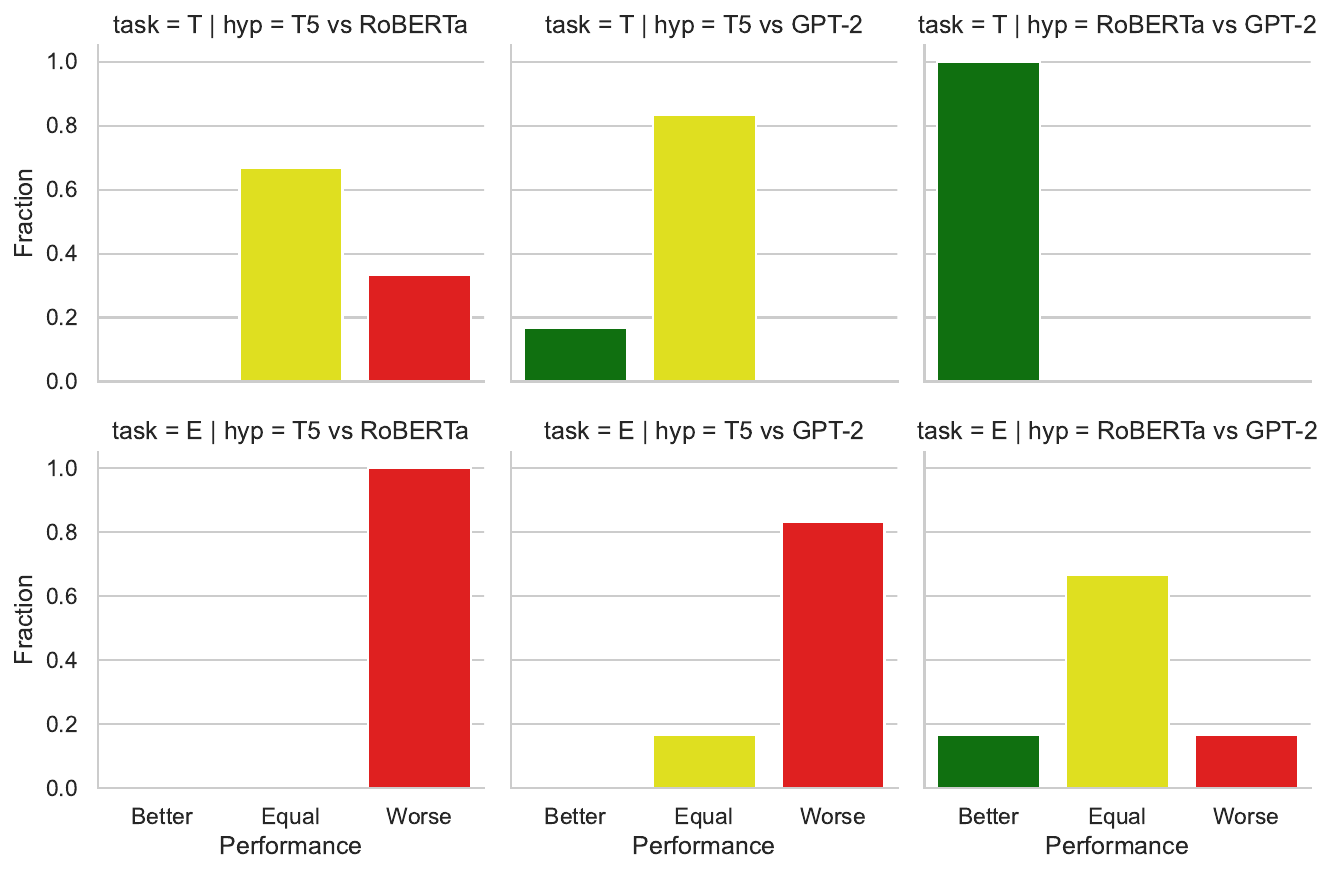}

    \caption{The proportion of \timex\ (upper panel) and \eventx\ (lower panel) datasets where a fine-tuned PLM $m_i$ is significantly better, worse, or almost equal to another model $m_j$ (\texttt{hyp = $m_i$ vs $m_j$}).}

    \label{fig:fine-tuned}
    \vspace{-8pt}
\end{figure}

\textbf{Fine-tuned Models.} While our main focus is prompted LLMs, we include a complementary analysis of fine-tuned pretrained models to examine \textit{consistency}, rather than absolute generalization performance. Following \citet{dutt2024investigating}, we train RoBERTa \citep{liu2019roberta}, T5 \citep{raffel2020exploring}, and GPT-2 \citep{radford2019language} on the base \timex\ and \eventx\ sequence labeling tasks, and evaluate on one dataset from each domain, length and adversarial dimension. Details are provided in Appendix \S~\ref{sec:fine-tuned-models}. Figure \ref{fig:fine-tuned} reports the proportion of generalization datasets where model $m_i$ is significantly better, worse, or almost equal to model $m_j$. Although all three PLMs perform poorly in some generalization dimensions, eg., \textit{Adv\_P} and \textit{T2C} (Table \ref{tab:finetuned_results}, \S \ref{sec:app-detailed-results}), their relative behavior is stable: RoBERTa consistently outperforms T5 and GPT-2 across all dimensions. This contrasts with prompted LLMs, where no configuration shows comparable cross-dimensional consistency.

\section{Related Work}

Prior work has studied NLP generalization from multiple perspectives and proposed methods to improve it across dimensions. \citep{hupkes2023taxonomy} provides a broad taxonomy of generalization, from which we focus on four key categories that cover most benchmark settings. Cross-task generalization captures a model's ability to transfer knowledge across tasks, where instruction-tuned LLMs trained on large-scale annotated data have shown strong performance \citep{zhang2026instruction}. Models such as GPT-3 \citep{brown2020language} and LLaMA2 \citep{touvron2023llama} further demonstrate zero-shot capabilities, though concerns about data contamination raise questions about whether these results reflect true generalization \citep{li2024task}. Beyond task transfer, prior work has also explored reasoning strategies as a lens for evaluating model behavior \citep{liu2024incomplete}. Within the dimensions we study, domain generalization has been widely explored \citep{ramponi2020neural, naik2022adapting}, while robustness is typically evaluated using adversarial perturbations, including both white-box and black-box attacks \citep{ebrahimi2018hotflip, jin2020bert, goyal2023survey}. However, many such methods produce unrealistic inputs and do not capture shortcut learning tendencies \citep{geirhos2020shortcut}. Compositional generalization has been studied across tasks such as machine translation, semantic parsing, and question answering \citep{dankers2022paradox, kim2020cogs, keysersmeasuring}, though systematic dataset construction via composition remains limited. Structural generalization, particularly length generalization, has received comparatively less attention, despite consistent evidence that models struggle in ``train-short, test-long'' settings \citep{press2108train, anil2022exploring, ruoss2023randomized}. \newline

In parallel, a large body of work has proposed methods to improve generalization. Domain adaptation approaches include gradual fine-tuning across domains and domain adversarial training \citep{xu2021gradual, wright2020transformer}. Robustness has been addressed through adversarial augmentation, loss regularization, and generative adversarial training \citep{li2019textbugger, goodfellow2015explaining, kang2018adventure}. For compositionality, earlier work relies on neuro-symbolic methods and meta-learning, while more recent approaches leverage structured prompting in LLMs \citep{guptaneural, lake2019compositional, press2023measuring}. Additionally, several studies explore the integration of inductive, deductive, and abductive reasoning to enhance model capabilities \citep{he2025idea, hu2025beyond, cao2026fundamental}. In contrast to work that improves generalization methods or evaluates broader temporal reasoning in LLMs \citep{fatemi2025test, bazaga2025learning, park2025chroknowledge}, we systematically assess how existing models generalize on the foundational extraction tasks of identifying time and event expressions. 

\section{Conclusion \& Future Work}
We present a systematic study of multidimensional generalization in prompted LLMs for time and event expression extraction, covering domain, robustness, compositional, and structural shifts. Our results show that base-task performance is not a reliable proxy for generalization across dimensions. Inductive reasoning strategies improve both base and generalization performance and yield stronger cross-dimensional correlations, but model size and architecture are not consistently predictive of generalization. In contrast, fine-tuned PLMs show more stable relative behavior across dimensions, though not necessarily stronger absolute performance. These findings suggest that reliable temporal extraction requires evaluation beyond in-domain performance and motivate our future work on models and prompting strategies that generalize more consistently across diverse distribution shifts.

\section*{Acknowledgments}

The authors acknowledge the Texas Advanced Computing Center (TACC) at The University of Texas at Austin for providing computational resources that contributed to the research results reported in this paper. More information is available at \url{https://www.tacc.utexas.edu}.

\section*{Limitations}
\label{sec:limitations}

Our study focuses on extracting time (\timex) and event (\eventx) expressions. These tasks are core primitives for temporal understanding, but they do not directly evaluate higher-level temporal reasoning tasks such as temporal relation inference, timeline construction, or temporal question answering. A second limitation is possible contamination from pretraining corpora, especially for widely used benchmark datasets. However, our main conclusions rely on relative degradation across distribution shifts, where models still show substantial drops, particularly under cross-domain and robustness-oriented settings. Thus, even if some base datasets were partially seen during pretraining, the generalization gaps remain informative.

Our model comparisons also have limits. We refer to all prompted generative models as LLMs for consistency, although some smaller models in our study may also be described as small language models (SLMs). The dense and MoE models differ not only in architecture, but also in active parameter count, pretraining data, and model generation. Therefore, our results should \textbf{not} be interpreted as isolating the causal effect of MoE architecture alone. Similarly, our prompting comparison reflects end-to-end prompting strategies as they are commonly used. Because reasoning style, prompt wording, and output format vary together, we cannot fully attribute performance differences to the reasoning strategy alone.

\bibliography{custom}

\newpage
\appendix

\section{ Appendix}
\label{sec:appendix}
\subsection{Additional Results}
\label{sec:app-exp-results}

\subsubsection{Base Task Performance}
\label{sec:app-ind-plus-vs-non-inductive}
\begin{figure*}[t!]
    \centering
    \includegraphics[width=0.75\textwidth]{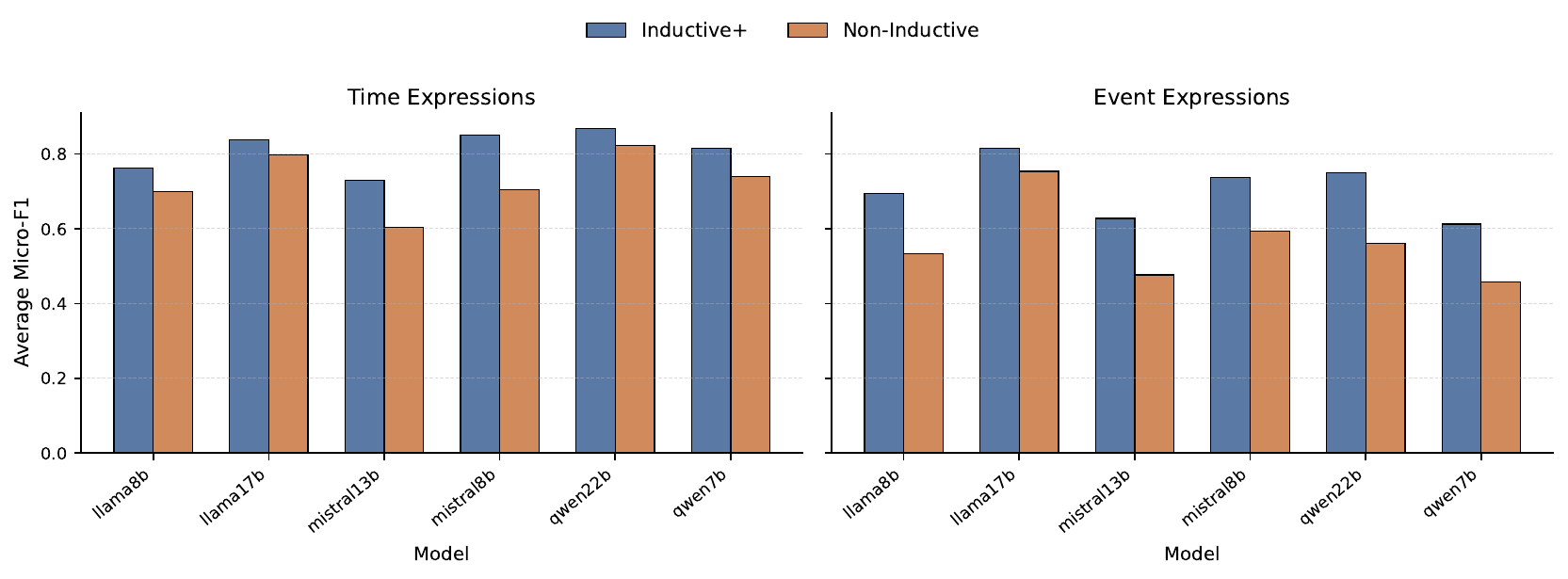}

    \caption{Base task performance in inductive+ vs.\ non-inductive strategies.}

    \label{fig:app-ind-plus-vs-non-inductive}
    \vspace{-8pt}
\end{figure*}

Figure~\ref{fig:app-ind-plus-vs-non-inductive} compares the average Base performance (Micro-F1) of each model under two setups: inductive+ and non-inductive. We observe that inductive+ configurations consistently achieve higher scores across nearly all models in both \timex\ and \eventx\ tasks. This suggests that incorporating inductive reasoning, through in-context examples, improves extraction performance on the base dataset.

\subsubsection{Correlation Across Generalization Dimensions}
\label{sec:app-corr-gen-dim}

\begin{figure*}[t!]
    \centering

    \begin{subfigure}[t]{0.49\textwidth}
        \centering
        \includegraphics[width=\linewidth]{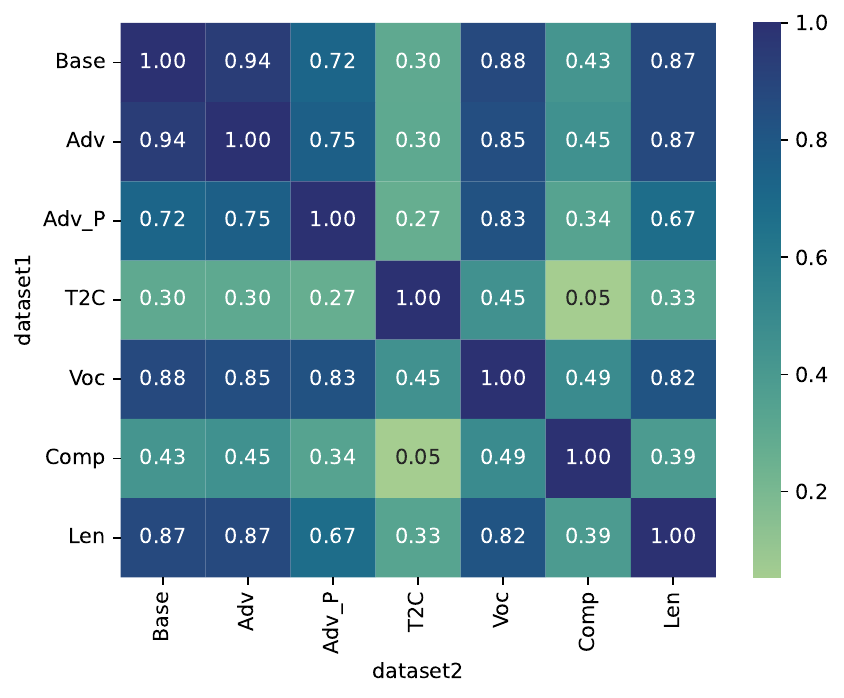}
        \caption{Inductive+ strategies}
        \label{fig:eventx-inductive-sr}
    \end{subfigure}
    \hfill
    \begin{subfigure}[t]{0.49\textwidth}
        \centering
        \includegraphics[width=\linewidth]{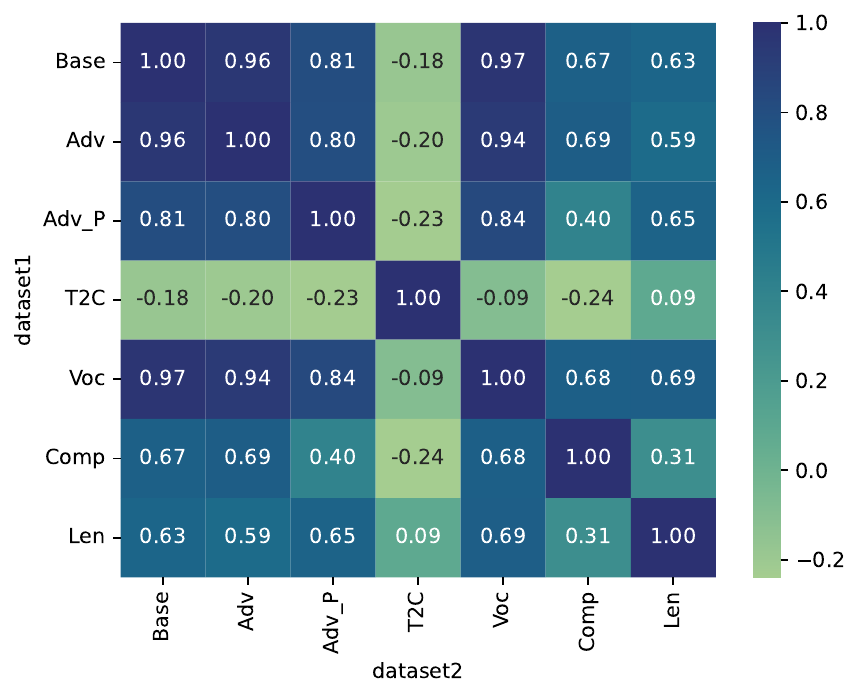}
        \caption{Non-inductive strategies}
        \label{fig:eventx-noninductive-sr}
    \end{subfigure}

    \caption{$\rho$ across generalization dimensions for event expression extraction in inductive+ and non-inductive strategies.}

    \label{fig:ind-vs-nonind-sr_2}
    \vspace{-8pt}
\end{figure*}

Figure~\ref{fig:ind-vs-nonind-sr_2} presents the configuration-wise rank correlation ($\rho$) across generalization dimensions for event extraction under inductive+ and non-inductive strategies. In the inductive+ setting (Figure~\ref{fig:eventx-inductive-sr}), we observe consistently strong correlations among Base, Adversarial, Vocabulary, and Length dimensions, indicating stable transfer of configuration rankings across these settings. In contrast, the compositional (Comp) and T2C dimensions exhibit notably weaker correlations, suggesting limited alignment with other dimensions. For non-inductive strategies (Figure~\ref{fig:eventx-noninductive-sr}), while Base, Adversarial, and Vocabulary remain moderately correlated, overall correlations are lower and more variable. Notably, in non-inductive strategies, the T2C dimension shows near-zero or negative correlations with most other dimensions, whereas in inductive+ setups, it shows much higher correlations.

\subsubsection{Effect of Reasoning Strategy}
\label{sec:app-reasoning-strategy}
\begin{figure*}[t!]
    \centering
    \includegraphics[width=0.9\textwidth]{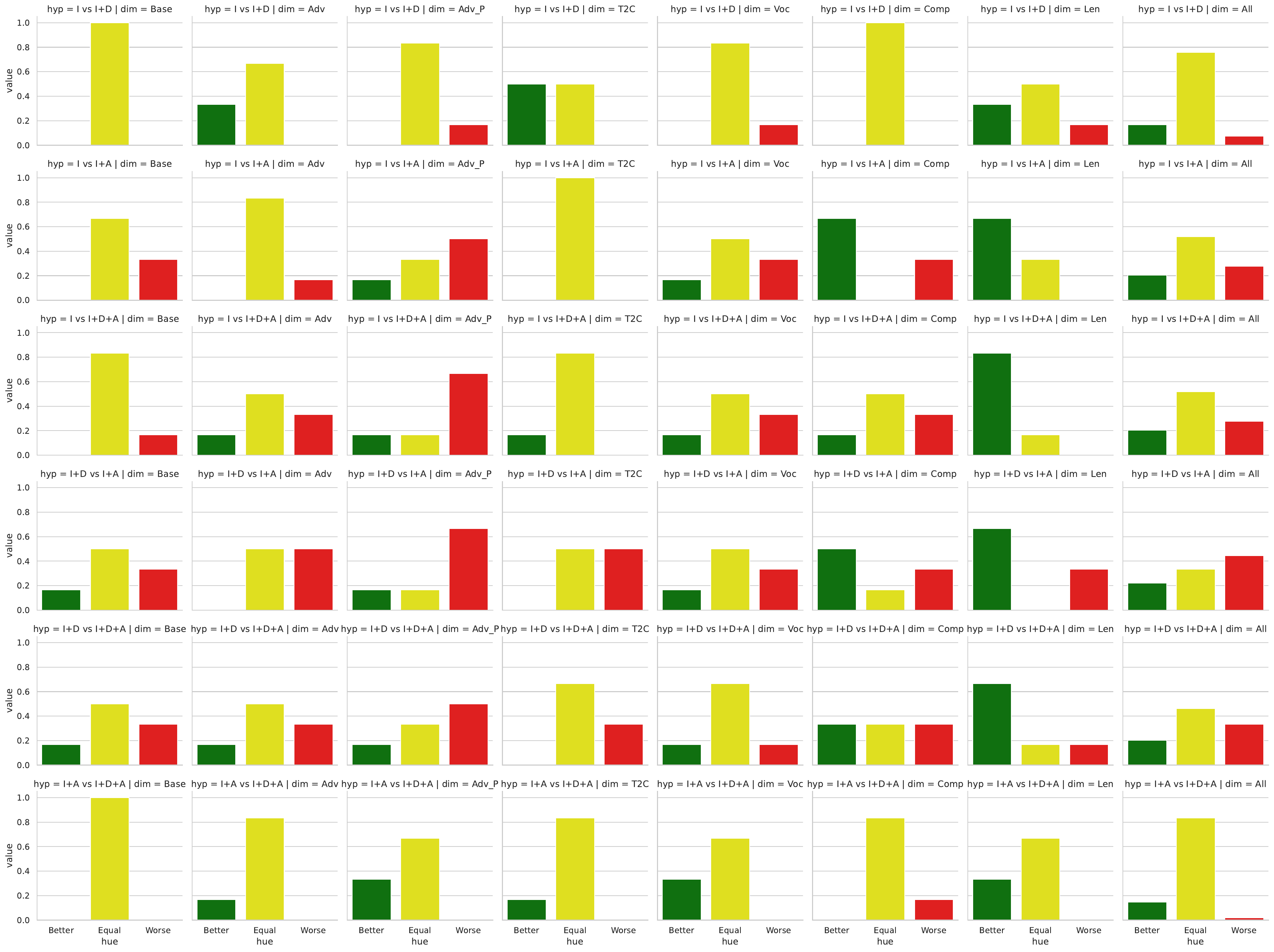}

    \caption{Percentage of models that perform \textbf{significantly} \textcolor{mygreen}{better}, \textcolor{myred}{worse}, or \textcolor{myyellow}{almost equal} for reasoning strategy $i$ vs.\ $j$, expressed as \texttt{Hyp = i vs j}, in each dimension (\texttt{Dim = x}) for \textit{time expression extraction tasks}. Each column corresponds to one dimension.}

    \label{fig:reasoning-strategy-time-all-dim}
    \vspace{-8pt}
\end{figure*}

Figure~\ref{fig:reasoning-strategy-time-all-dim} shows the percentage of configurations where one reasoning strategy outperforms, underperforms, or remains indistinguishable from another across generalization dimensions for time extraction. Most comparisons fall into the \textit{equal} category, suggesting similar performance across strategies. Inductive-based methods provide small but consistent gains in Base, Adversarial, and Vocabulary settings, while differences become more variable in compositional and cross-domain (T2C) dimensions. Moreover, combining all reasoning types ($I{+}D{+}A$) does not consistently improve performance, indicating diminishing returns from added complexity.

\begin{figure*}[t!]
    \centering
    \includegraphics[width=\textwidth]{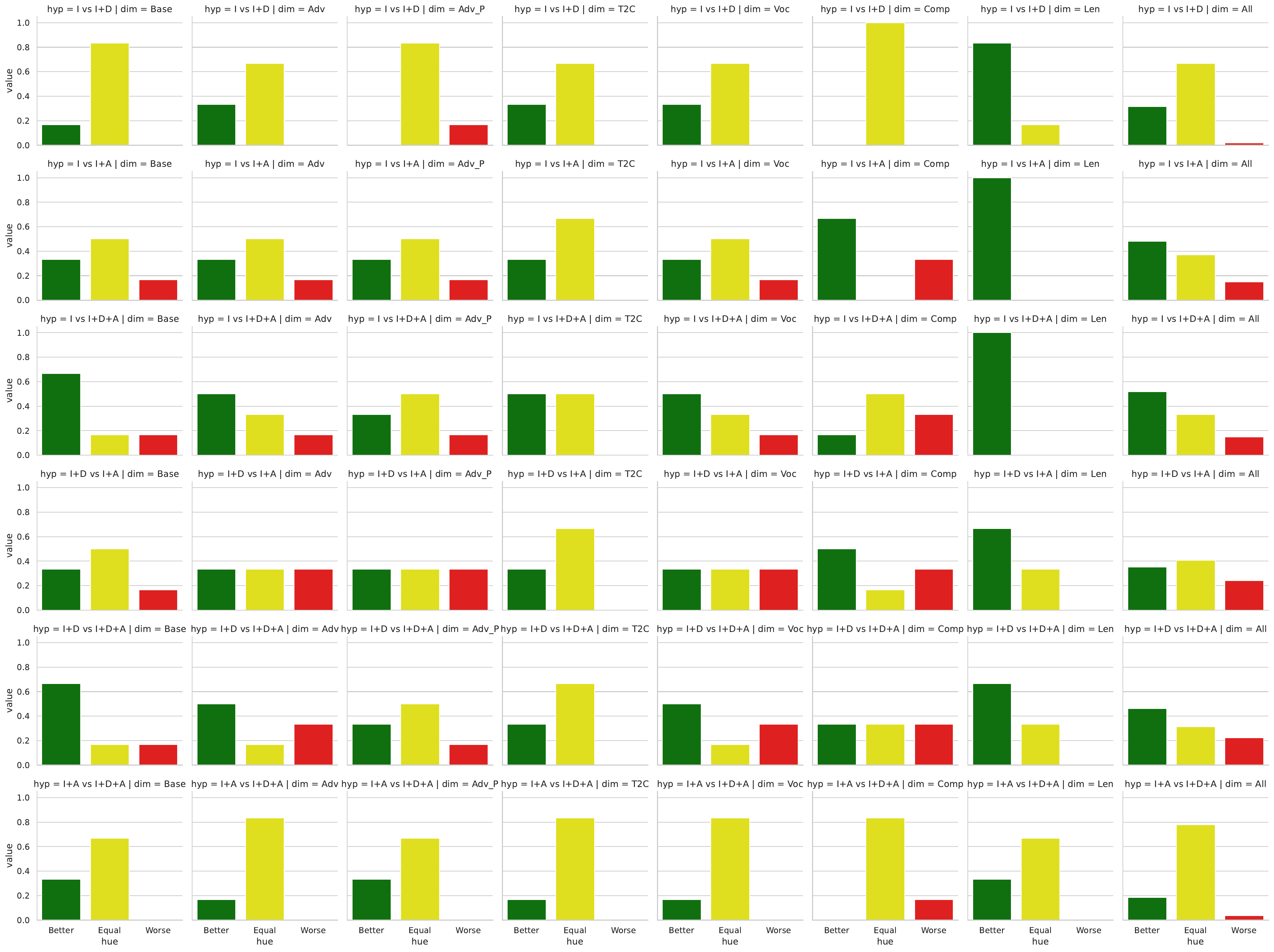}

    \caption{Percentage of models that perform \textbf{significantly} \textcolor{mygreen}{better}, \textcolor{myred}{worse}, or \textcolor{myyellow}{almost equal} for reasoning strategy $i$ vs.\ $j$, expressed as \texttt{Hyp = i vs j}, in each dimension (\texttt{Dim = x}) for \textit{event expression extraction tasks}. Each column corresponds to one dimension.}

    \label{fig:reasoning-strategy-event-all-dim}
    \vspace{-8pt}
\end{figure*}

Figure~\ref{fig:reasoning-strategy-event-all-dim} summarizes the same comparison for event extraction. Again, most strategy pairs perform similarly, with inductive variants remaining competitive but less consistently beneficial than in time extraction. Variability increases in compositional and cross-domain settings, where no single strategy consistently dominates.

\subsection{Model Training Details}
\label{sec:fine-tuned-models}

\paragraph{Overview.}
We fine-tune three pretrained transformer models—RoBERTa, T5, and GPT-2 on two sequence labeling tasks: temporal expression extraction (timex) and event trigger extraction (eventx). Each model is trained independently for each task, resulting in six fine-tuned models. All models are adapted for token classification by adding a linear classification head over token representations. Training is performed using the HuggingFace Transformers library with the Trainer API and the AdamW optimizer.

\paragraph{Task Formulation.}
Both tasks are formulated as named entity recognition (NER) problems in CoNLL format, where each line contains a token and its corresponding label, and sentences are separated by blank lines. 
We convert the original annotations into a BIO tagging format. Temporal expressions and event mentions are mapped to token-level labels, where each span is assigned a beginning (B) and inside (I) tag, and all other tokens are labeled as outside (O). Specifically:
\begin{itemize}
    \item \textbf{timex:} B-T, I-T, O
    \item \textbf{eventx:} B-E, I-E, O
\end{itemize}

\paragraph{Dataset.}
We train all models on the base temporal expression and event extraction datasets. Each task is represented in CoNLL format with predefined train and test splits. The label encoder is fit on the union of training and test labels to ensure consistent label mappings across splits.

\paragraph{Models.}
We evaluate three transformer architectures:
\begin{itemize}
    \item \textbf{RoBERTa (roberta-base):} encoder-only model with 12 layers and hidden size 768, implemented using \texttt{RobertaForTokenClassification}. Tokenization uses \texttt{add\_prefix\_space=True}.
    \item \textbf{T5 (flan-t5-small):} encoder-decoder model adapted for token classification using \texttt{T5ForTokenClassification}, configured to operate as an encoder for this task.
    \item \textbf{GPT-2 (gpt2):} decoder-only model using \texttt{GPT2ForTokenClassification}. The EOS token is used as the padding token, and embeddings are resized using \texttt{resize\_token\_embeddings()}.
\end{itemize}

\paragraph{Tokenization and Label Alignment.}
All models use subword tokenization. Word-level labels are aligned to subword tokens by assigning the original label to all subword pieces of a word. Special tokens and padding tokens are assigned an ignore index of $-100$. All sequences use the maximum context length supported by each model.

\paragraph{Training Configuration.}
All models are trained with identical hyperparameters:
\begin{itemize}
    \item Optimizer: AdamW
    \item Learning rate: $2 \times 10^{-5}$
    \item Weight decay: 0.01
    \item Warmup steps: 500 (linear schedule with decay)
    \item Epochs: 3
    \item Train batch size: 16
    \item Eval batch size: 32
    \item Max sequence length: Full model context length
\end{itemize}

Model checkpoints are saved and evaluated every 250 steps. Early stopping is applied with a patience of 3 evaluations. Model selection is based on entity-level F1 score. All model parameters are updated during training (full fine-tuning); no parameter-efficient fine-tuning methods (e.g., LoRA or adapters) are used.

\begin{figure}[t!]
\centering

\begin{tcolorbox}[
    width=\columnwidth,
    colback=gray!5,
    colframe=black,
    boxrule=0.5pt,
    arc=2pt,
    left=6pt,
    right=6pt,
    top=6pt,
    bottom=6pt
]
\footnotesize

\textbf{System Prompt:}

You are an expert system for time expression extraction in natural language text.

\vspace{0.5em}
\textbf{Task Definition:}

You are given annotated examples and a new input text.

Your task is to extract all explicit time expressions from the new input text by generalizing from the examples. Time expressions may include dates, times, durations, frequencies, or other grounded temporal references explicitly stated in the text.

\vspace{0.5em}
\textbf{REASONING TYPE DEFINITION:}

You must use inductive reasoning, defined as follows:

Infer which spans qualify as time expressions based on the annotated examples

Learn the typical lexical forms and boundary conventions of these expressions

Apply the inferred patterns consistently to unseen input text

You should not rely on external knowledge or assumptions beyond what can be inferred from the examples.

\vspace{0.5em}
\textbf{OUTPUT FORMAT:}

Return ONLY the following:

A valid JSON array of strings

Each string is one extracted time expression

No duplicates

Ordered by first appearance in the text

No additional text, explanation, or formatting

\vspace{0.5em}
\textbf{Examples:} \\
\textit{(Few-shot annotated examples)}

\vspace{0.5em}
\textbf{Input:} \\
\textit{(New context)}

\end{tcolorbox}

\caption{Inductive prompting template used for time expression extraction.}
\label{fig:inductive-prompt}
\vspace{-8pt}

\end{figure}

\begin{figure}[t!]
\centering

\begin{tcolorbox}[
    width=\columnwidth,
    colback=gray!5,
    colframe=black,
    boxrule=0.5pt,
    arc=2pt,
    left=6pt,
    right=6pt,
    top=6pt,
    bottom=6pt
]
\footnotesize

\textbf{System Prompt:}

You are an expert system for time expression extraction in natural language text.

\vspace{0.5em}
\textbf{Task Definition:}

You are given a new input text.

Your task is to extract all explicit time expressions from the text by applying predefined rules. Time expressions may include dates, times, durations, frequencies, or other grounded temporal references explicitly stated in the text.

A span qualifies as a time expression if it explicitly denotes:

A calendar date

A clock time

A duration

A frequency or recurrence

Another grounded temporal reference

\vspace{0.5em}
\textbf{REASONING TYPE DEFINITION:}

You must use deductive reasoning, defined as follows:

Apply the given definition of time expressions directly to the input text

Select only spans that strictly satisfy the stated criteria

Do not infer new rules or adapt based on the input text

You should rely only on the provided rules and not on examples or learned patterns.

\vspace{0.5em}
\textbf{OUTPUT FORMAT:}

Return ONLY the following:

A valid JSON array of strings

Each string is one extracted time expression

No duplicates

Ordered by first appearance in the text

No additional text, explanation, or formatting

\vspace{0.5em}
\textbf{Input:} \\
\textit{(New context)}

\end{tcolorbox}

\caption{Deductive prompting template used for time expression extraction.}
\label{fig:deductive-prompt}
\vspace{-8pt}

\end{figure}

\begin{figure}[t!]
\centering

\begin{tcolorbox}[
    width=\columnwidth,
    colback=gray!5,
    colframe=black,
    boxrule=0.5pt,
    arc=2pt,
    left=6pt,
    right=6pt,
    top=6pt,
    bottom=6pt
]
\footnotesize

\textbf{System Prompt:}

You are an expert system for time expression extraction in natural language text.

\vspace{0.5em}
\textbf{Task Definition:}

You are given a new input text.

Your task is to extract all explicit time expressions by using abductive reasoning. Time expressions may include dates, times, durations, frequencies, or other grounded temporal references explicitly stated in the text.

\vspace{0.5em}
\textbf{REASONING TYPE DEFINITION:}

You must use abductive reasoning and think step by step, defined as follows:

Generate candidate spans that could plausibly express temporal information

Reason about how each candidate explains the timing, duration, or recurrence of events in the text

Select the spans that best account for the author’s intended temporal meaning

Prefer the most complete and non-overlapping spans

\vspace{0.5em}
\textbf{OUTPUT FORMAT:}

Return ONLY a valid JSON list of dictionaries.

Each dictionary must contain the following fields:

\texttt{"text"}: the original input text

\texttt{"time\_expressions"}: a list of extracted time expression strings (verbatim from the text). If no time expressions are present, \texttt{"time\_expressions"} must be an empty list ([])

\texttt{"reasoning\_steps"}: a list of strings explaining the abductive reasoning process

\vspace{0.5em}

\textbf{Example Output:}

\begin{Verbatim}[fontsize=\tiny, breaklines=true, breaksymbolleft={}]
[
  {
    "text": "The conference started on April 9, 1997 and lasted for three days.",
    "time_expressions": ["April 9, 1997", "three days"],
    "reasoning_steps": [
      "Identified a calendar date.",
      "Identified a duration phrase.",
      "Selected both spans."
    ]
  }
]
\end{Verbatim}

\vspace{0.5em}
\textbf{Input:} \\
\textit{(New context)}

\end{tcolorbox}

\caption{Abductive prompting template used for time expression extraction.}
\label{fig:abductive-prompt}
\vspace{-8pt}

\end{figure}

\begin{figure}[t!]
\centering

\begin{tcolorbox}[
    width=\columnwidth,
    colback=gray!5,
    colframe=black,
    boxrule=0.5pt,
    arc=2pt,
    left=6pt,
    right=6pt,
    top=6pt,
    bottom=6pt
]

\footnotesize

\textbf{System Prompt:}

You are an expert system for time expression extraction in natural language text.

\vspace{0.5em}
\textbf{Task Definition:}

You are given annotated examples and a new input text.

Your task is to extract all explicit time expressions from the new input text. Time expressions may include dates, times, durations, frequencies, or other grounded temporal references explicitly stated in the text.

\vspace{0.5em}
\textbf{REASONING TYPE DEFINITION:}

You must use inductive, deductive, and abductive reasoning, and think step by step, defined as follows:

Inductively infer from the annotated examples which spans qualify as time expressions, including their typical lexical forms and boundary conventions.

Deductively apply the inferred criteria to the new input text to identify candidate spans that satisfy the definition of a time expression.

Abductively select the spans that best explain the temporal meaning intended by the author, preferring the most complete and non-overlapping spans.

You should not rely on external knowledge or assumptions beyond what can be inferred from the examples and the induced criteria.

\vspace{0.5em}
\textbf{OUTPUT FORMAT:}

Return ONLY a valid JSON list of dictionaries.

Each dictionary must contain the following fields:

\texttt{"text"}: the original input text

\texttt{"time\_expressions"}: a list of extracted time expression strings (verbatim from the text). If no time expressions are present, \texttt{"time\_expressions"} must be an empty list (\texttt{[]})

\texttt{"reasoning\_steps"}: a list of strings explaining the reasoning process

\vspace{0.5em}

\textbf{Example Output:}

\begin{Verbatim}[fontsize=\tiny, breaklines=true, breaksymbolleft={}]
[
  {
    "text": "The conference started on April 9, 1997 and lasted for three days.",
    "time_expressions": ["April 9, 1997", "three days"],
    "reasoning_steps": [
      "Identified a calendar date.",
      "Identified a duration phrase.",
      "Selected both spans."
    ]
  }
]
\end{Verbatim}

\vspace{0.5em}
\textbf{Input:} \\
\textit{(New context)}

\end{tcolorbox}

\caption{Inductive, deductive, and abductive prompting template used for time expression extraction.}
\label{fig:ida-prompt}
\vspace{-8pt}

\end{figure}

\subsection{Prompting Templates for Reasoning Strategies}
\label{sec:appendix_prompts}

In this section, we provide the prompting templates for the three reasoning strategies: inductive (I), deductive (D), and abductive (A)—for the temporal expression extraction task. The same overall prompt structure is used for both temporal and event extraction tasks, although the task definitions and extraction criteria differ for each task. These reasoning prompts are also used consistently across all generalization settings (e.g., domain, compositional, robustness, and structural), ensuring that differences in performance arise from the evaluation conditions rather than changes in prompt design. 

\subsubsection{Inductive Prompt}
\label{sec:Inductive}
Inductive reasoning in our temporal extraction task is formulated as an example-driven span identification process, where the model is provided with annotated instances of valid time expressions, as illustrated in Figure~\ref{fig:inductive-prompt} . By observing these examples, the model learns the typical lexical patterns and boundary conventions associated with temporal spans, such as dates, durations, and frequencies. It then generalizes these patterns to extract similar expressions from unseen input text.

\subsubsection{Deductive Prompt}
\label{sec:Deductive}
Deductive reasoning in our temporal extraction task is formulated as a rule-based span selection process, where the definition of a valid time expression is specified explicitly in the prompt (Figure~\ref{fig:deductive-prompt}). The model is required to identify spans that directly correspond to grounded temporal references—such as dates, times, durations, and frequencies by strictly applying these criteria to the input text. This setup removes reliance on example-based learning and constrains the model to make extraction decisions based solely on the provided definition, ensuring consistent and precise identification of temporal spans.

\subsubsection{Abductive Prompt}
\label{sec:Abductive}

In the temporal extraction task, abductive reasoning is implemented as a hypothesis-driven process in which the model generates candidate temporal spans and selects those that best explain the temporal structure of the text (Figure~\ref{fig:abductive-prompt}). The model is also required to generate explicit explanations describing its selection of the final time expressions. This allows the model to incorporate contextual cues and resolve ambiguities in identifying time expressions. For prompting settings involving abductive reasoning, we additionally provide one structured example to guide smaller models in following the required output format.

\subsubsection{Inductive + Deductive + Abductive Prompt}
\label{sec:IDA}
Inductive, deductive, and abductive reasoning in our temporal extraction task is formulated as a combined example-guided and hypothesis-driven span identification process (Figure~\ref{fig:ida-prompt}). The model first learns from annotated examples which kinds of spans typically function as time expressions and what their boundary conventions look like. It then applies these inferred criteria to the new input text to identify candidate temporal spans. Finally, it selects the spans that best explain the temporal meaning of the text, favoring complete and non-overlapping expressions, while also providing brief reasoning steps for its decisions.

\subsection{Datasets}
\label{sec:app-datasets}

The original document-level \timex\ and \eventx\ datasets derived from the TimeBank corpus contain 2.12K training instances, 265 validation instances, and 266 test instances each. From these document-level datasets, we construct sentence-level variants for both tasks. The sentence-level \timex\ dataset contains 22.7K training instances, 2.67K validation instances, and 400 test instances. Similarly, the sentence-level \eventx\ dataset contains 23.2K training instances, 3.16K validation instances, and 400 test instances.

The adversarial datasets (\textit{Adv} and \textit{Adv\_P}) are generated from the sentence-level test sets for both \timex\ and \eventx. The vocabulary-shift datasets (\textit{Voc}) are also constructed from the same sentence-level test sets. 

For structural generalization (\textit{Len}), we use 80 document-level instances from the original TimeBank datasets for evaluation. These instances have a minimum sequence length of 63 tokens, a maximum sequence length of 1134 tokens, and an average sequence length of 369.98 tokens. Among the 80 instances, 73 contain fewer than 512 tokens, which is the smallest context window among the evaluated PLM models. This helps ensure that our structural generalization results are not affected by the documents being truncated. For all prompted models, the context length exceeds the maximum sequence length.

For compositional generalization (\textit{Comp}), we select sentence-level instances from the original document-level corpus that contain both time and event expressions. The resulting evaluation set contains 400 instances.

\subsubsection{Label and Vocabulary Shift in T2C}
\label{sec:app-t2c-vocab-shift}
For the clinical domain-shift datasets (\textit{T2C}) derived from the 2012 i2b2 Clinical Temporal Relations corpus, the clinical \timex\ dataset contains 1.32K training instances, 1.35K validation instances, and 400 test instances. The clinical \eventx\ dataset contains 5.56K training instances, 1.35K validation instances, and 400 test instances.

The statistics in Table~\ref{tab:top_time_event_expressions} highlight substantial distributional differences between the original TimeBank labels (news-oriented corpus) and the clinical \textit{T2C} labels, which reflect the specialized language and domain-specific characteristics of clinical narratives. 

\begin{table*}[t]
\centering
\small
\caption{Top 10 most frequent time and event expressions across the Base (\textit{Base}) and clinical (\textit{T2C}) datasets.}
\label{tab:top_time_event_expressions}
\begin{tabular}{c p{2.0cm} p{2.0cm} p{2.0cm} p{2.0cm}}
\toprule
\textbf{Rank} &
\textbf{Base \timex} &
\textbf{\textit{T2C} \timex} &
\textbf{Base \eventx} &
\textbf{\textit{T2C} \eventx} \\
\midrule

1 & friday (25) & the time (9) & said (187) & admission (279) \\
2 & tuesday (25) & the day (9) & talks (27) & discharge (237) \\
3 & monday (24) & that time (5) & told (20) & showed (139) \\
4 & wednesday (17) & the night (5) & take (20) & admitted (122) \\
5 & thursday (17) & day of life two (5) & meeting (18) & transferred (116) \\
6 & now (14) & one (4) & war (14) & revealed (74) \\
7 & sunday (12) & the time of discharge (4) & killed (13) & discharged (59) \\
8 & saturday (12) & daily (4) & expected (13) & presented (53) \\
9 & this year (7) & few days (4) & set (13) & the floor (43) \\
10 & five years (6) & bid (4) & deal (13) & intubated (42) \\

\bottomrule
\end{tabular}
\end{table*}




\subsubsection{Time Expression extraction}
\label{sec:datasets}

We evaluate time expression extraction under six different settings: \textit{Base}, \textit{Adversarial}, \textit{Domain}, \textit{Domain-Vocab}, \textit{Compositional}, and \textit{Structural}. The base setting uses sentence-level examples from TimeBank, while the adversarial setting adds perturbations to test robustness. The domain and domain-vocabulary settings measure generalization to clinical and modified vocabulary data, and the compositional setting requires extracting both time expressions and event triggers together. The structural setting uses longer document-level inputs with multiple temporal references and more complex context.  Figure~\ref{fig:timex_examples} shows example instances from each setting. These settings provide a comprehensive evaluation framework for analyzing temporal extraction generalization across diverse conditions.

\begin{figure*}[!t]
\centering

\begin{minipage}[t]{0.48\textwidth}
\begin{tcolorbox}[
    width=\linewidth,
    colback=gray!15,
    colframe=black,
    boxrule=0.5pt,
    arc=0pt,
    left=5pt,
    right=5pt,
    top=5pt,
    bottom=5pt
]
\scriptsize

\begin{center}
\hypertarget{box:base}{}
\textbf{\textcolor{basecolor}{Base}}
\end{center}

\textbf{Text:} MHB technicians said they started the surveillance mission Thursday to monitor the body temperature of all Macao residents and visitors.\\

\textbf{Extracted span:} Thursday

\end{tcolorbox}
\end{minipage}
\hfill
\begin{minipage}[t]{0.48\textwidth}
\begin{tcolorbox}[
    width=\linewidth,
    colback=gray!15,
    colframe=black,
    boxrule=0.5pt,
    arc=0pt,
    left=5pt,
    right=5pt,
    top=5pt,
    bottom=5pt
]
\scriptsize

\begin{center}
\hypertarget{box:adv}{}
\textbf{\textcolor{advcolor}{Adversarial}}
\end{center}

\textbf{Text:} MHB technicians \textcolor{red}{tell} they \textcolor{red}{launches} the \textcolor{red}{monitor remit} Thursday to \textcolor{red}{observer} the \textcolor{red}{entity warmth} of all Macao \textcolor{red}{reside} and \textcolor{red}{traveler}.\\

\textbf{Extracted span:} Thursday

\end{tcolorbox}
\end{minipage}

\vspace{0.3cm}

\begin{minipage}[t]{0.48\textwidth}
\begin{tcolorbox}[
    width=\linewidth,
    colback=gray!15,
    colframe=black,
    boxrule=0.5pt,
    arc=0pt,
    left=5pt,
    right=5pt,
    top=5pt,
    bottom=5pt
]
\scriptsize

\begin{center}
\hypertarget{box:domain}{}
\textbf{\textcolor{domaincolor}{Domain(\textit{T2C})}}
\end{center}

\textbf{Text:} Pregnancy was IUI assisted for di-di twin gestation and was relatively uncomplicated until mother presented on 12-15, with spontaneous rupture of membranes.\\

\textbf{Extracted span:} 12-15

\end{tcolorbox}
\end{minipage}
\hfill
\begin{minipage}[t]{0.48\textwidth}
\begin{tcolorbox}[
    width=\linewidth,
    colback=gray!15,
    colframe=black,
    boxrule=0.5pt,
    arc=0pt,
    left=5pt,
    right=5pt,
    top=5pt,
    bottom=5pt
]
\scriptsize

\begin{center}
\hypertarget{box:domainvocab}{}
\textbf{\textcolor{vocabcolor}{Domain-Vocab}}
\end{center}

\textbf{Text:} MHB technicians said they started the surveillance mission Thursday to monitor the body temperature of all \dv{Myfiu} residents and visitors.\\

\textbf{Extracted span:} Thursday

\end{tcolorbox}
\end{minipage}

\vspace{0.3cm}

\begin{minipage}[t]{0.98\textwidth}
\begin{tcolorbox}[
    width=\linewidth,
    colback=gray!15,
    colframe=black,
    boxrule=0.5pt,
    arc=0pt,
    left=5pt,
    right=5pt,
    top=5pt,
    bottom=5pt
]
\scriptsize

\begin{center}
\hypertarget{box:comp}{}
\textbf{\textcolor{compcolor}{Compositional}}
\end{center}

\textbf{Text:} The longtime overheating then led to a blowout and the cold air went into the T101 aniline processing tower, which was run under negative pressure and triggered the blast at about 1:45 p.m., and other blasts happened afterwards.\\

\textbf{Extracted time and event spans:} 1:45 p.m.; overheating; led; blowout; went; run; triggered; blast; blasts; happened

\end{tcolorbox}
\end{minipage}

\vspace{0.3cm}

\begin{minipage}[t]{0.98\textwidth}
\begin{tcolorbox}[
    width=\linewidth,
    colback=gray!15,
    colframe=black,
    boxrule=0.5pt,
    arc=0pt,
    left=5pt,
    right=5pt,
    top=5pt,
    bottom=5pt
]
\scriptsize

\begin{center}
\hypertarget{box:struct}{}
\textbf{\textcolor{lengthcolor}{Structural}}
\end{center}

\textbf{Text:} XIN\_ENG\_20061129.0012 WASHINGTON, Nov. 29, 2006 (Xinhua) Clippers, Dunleavy reportedly reach four-year deal. The Los Angeles Clippers and coach Mike Dunleavy have reached a four-year contract extension worth more than 21 million U.S. dollars, according to the Los Angeles Times Tuesday. The team and Dunleavy's agent have worked out a deal that will pay him more than 5.3 million U.S. dollars a season. The Clippers have not made the deal public because bonuses tied to future playoff appearances were still being worked out, the Times said. The Clippers improved dramatically under Dunleavy last season, posting their best win total since the franchise's glory days as the Buffalo Braves in the mid-1970s. The team reached the playoffs for the first time in nine seasons and came within one win of reaching the Western Conference final. The Times reported the offer is close to the 6 million U.S. a year Dunleavy's agent initially proposed and would make him among the highest-paid coaches in the NBA. Dunleavy, the NBA's 1999 coach of the year, currently earns 2.5 million U.S. dollars a season, and his deal expires at the end of this season.\\

\textbf{Extracted spans:} Nov. 29, 2006; Times Tuesday; last season; mid-1970s; nine seasons; the year; season; the end of this season

\end{tcolorbox}
\end{minipage}

\caption{Example instances from the six dataset settings used in our time expression extraction experiments: base, adversarial, domain, domain-vocab, compositional, and structural. In the adversarial example, perturbed tokens are highlighted in red.}

\label{fig:timex_examples}
\vspace{-8pt}

\end{figure*}

\newpage
\subsubsection{Event Expression extraction}
\label{sec:datasets2}
Unlike time expression extraction, the event extraction task focuses on identifying event trigger words that describe actions, occurrences, or state changes in text. We evaluate the same six settings to analyze how well models generalize under perturbations, domain shifts, vocabulary changes, compositional reasoning, and longer contextual inputs. Figure~\ref{fig:event_examples} presents representative examples from each dataset setting, with adversarial perturbations highlighted in red.

\begin{figure*}[!t]
\centering

\begin{minipage}[t]{0.48\textwidth}
\begin{tcolorbox}[
    width=\linewidth,
    colback=gray!15,
    colframe=black,
    boxrule=0.5pt,
    arc=0pt,
    left=5pt,
    right=5pt,
    top=5pt,
    bottom=5pt
]
\scriptsize
\begin{center}
\hypertarget{box:base-event}{}
\textbf{\textcolor{basecolor}{Base}}
\end{center}
\textbf{Text:} The typhoon, which hit the eastern region of Bicol before dawn Thursday, packing maximum winds of 150 kilometers (93 miles) per hour, knocked out electrical services, telephone lines and water supply, further hampering rescue and relief efforts.\\
\textbf{Extracted spans:} hit; packing; knocked; hampering
\end{tcolorbox}
\end{minipage}
\hfill
\begin{minipage}[t]{0.48\textwidth}
\begin{tcolorbox}[
    width=\linewidth,
    colback=gray!15,
    colframe=black,
    boxrule=0.5pt,
    arc=0pt,
    left=5pt,
    right=5pt,
    top=5pt,
    bottom=5pt
]
\scriptsize
\begin{center}
\hypertarget{box:adv-event}{}
\textbf{\textcolor{advcolor}{Adversarial}}
\end{center}
\textbf{Text:} The typhoon, which hit the eastern region of Bicol before dawn Thursday, packing \textcolor{red}{maimum} winds of 150 kilometers (93 miles) per hour, knocked out electrical services, telephone lines and \textcolor{red}{kater} supply, further hampering rescue and relief efforts.\\
\textbf{Extracted spans:} hit; packing; knocked; hampering
\end{tcolorbox}
\end{minipage}

\vspace{0.3cm}

\begin{minipage}[t]{0.48\textwidth}
\begin{tcolorbox}[
    width=\linewidth,
    colback=gray!15,
    colframe=black,
    boxrule=0.5pt,
    arc=0pt,
    left=5pt,
    right=5pt,
    top=5pt,
    bottom=5pt
]
\scriptsize
\begin{center}
\hypertarget{box:domain-event}{}
\textbf{\textcolor{domaincolor}{Domain(\textit{T2C})}}
\end{center}
\textbf{Text:} The patient was reportedly alert at scene, and then became unresponsive in route to the hospital.\\
\textbf{Extracted spans:} alert; unresponsive; the hospital
\end{tcolorbox}
\end{minipage}
\hfill
\begin{minipage}[t]{0.48\textwidth}
\begin{tcolorbox}[
    width=\linewidth,
    colback=gray!15,
    colframe=black,
    boxrule=0.5pt,
    arc=0pt,
    left=5pt,
    right=5pt,
    top=5pt,
    bottom=5pt
]
\scriptsize
\begin{center}
\hypertarget{box:domainvocab-event}{}
\textbf{\textcolor{vocabcolor}{Domain-Vocab}}
\end{center}
\textbf{Text:} The typhoon, which hit the eastern region of \dv{Tcofa} before dawn Thursday, packing maximum winds of 150 kilometers (93 miles) per hour, knocked out electrical services, telephone lines and water supply, further hampering rescue and relief efforts.\\
\textbf{Extracted spans:} hit; packing; knocked; hampering
\end{tcolorbox}
\end{minipage}

\vspace{0.3cm}

\begin{minipage}[t]{0.98\textwidth}
\begin{tcolorbox}[
    width=\linewidth,
    colback=gray!15,
    colframe=black,
    boxrule=0.5pt,
    arc=0pt,
    left=5pt,
    right=5pt,
    top=5pt,
    bottom=5pt
]
\scriptsize
\begin{center}
\hypertarget{box:comp-event}{}
\textbf{\textcolor{compcolor}{Compositional}}
\end{center}
\textbf{Text:} The longtime overheating then led to a blowout and the cold air went into the T101 aniline processing tower, which was run under negative pressure and triggered the blast at about 1:45 p.m., and other blasts happened afterwards.\\
\textbf{Extracted time and event spans:} 1:45 p.m.; overheating; led; blowout; went; run; triggered; blast; blasts; happened
\end{tcolorbox}
\end{minipage}

\vspace{0.3cm}

\begin{minipage}[t]{0.98\textwidth}
\begin{tcolorbox}[
    width=\linewidth,
    colback=gray!15,
    colframe=black,
    boxrule=0.5pt,
    arc=0pt,
    left=5pt,
    right=5pt,
    top=5pt,
    bottom=5pt
]
\scriptsize
\begin{center}
\hypertarget{box:struct-event}{}
\textbf{\textcolor{lengthcolor}{Structural}}
\end{center}
\textbf{Text:} XIN\_ENG\_20061129.0012 WASHINGTON, Nov. 29, 2006 (Xinhua) Clippers, Dunleavy reportedly reach four-year deal. The Los Angeles Clippers and coach Mike Dunleavy have reached a four-year contract extension worth more than 21 million U.S. dollars, according to the Los Angeles Times Tuesday. The team and Dunleavy's agent have worked out a deal that will pay him more than 5.3 million U.S. doallars a season. The Clippers have not made the deal public because bonuses tied to future playoff appearances were still being worked out, the Times said. The Clippers improved dramatically under Dunleavy last season, posting their best win total since the franchise's glory days as the Buffalo Braves in the mid-1970s. The team reached the playoffs for the first time in nine seasons and came within one win of reaching the Western Conference final. The Times reported the offer is close to the 6 million U.S. a year Dunleavy's agent initially proposed and would make him among the highest-paid coaches in the NBA. Dunleavy, the NBA's 1999 coach of the year, currently earns 2.5 million U.S. dollars a season, and his deal expires at the end of this season.\\
\textbf{Extracted spans:} reached; according; worked; deal; pay; made; tied; worked; said; improved; posting; reached; came; reaching; reported; offer; proposed; make; earns; deal; expires
\end{tcolorbox}
\end{minipage}

\caption{Example instances from the six dataset settings used in our event extraction experiments: base, adversarial, domain, domain-vocab, compositional, and structural. In the adversarial example, perturbed tokens are highlighted in red.}
\label{fig:event_examples}
\vspace{-8pt}

\end{figure*}

\subsection{Detailed Results}
\label{sec:app-detailed-results}

This section presents the complete experimental results across all extraction tasks, model families, reasoning strategies, and generalization settings. We include detailed results in Tables~\ref{tab:moe_results}--\ref{tab:finetuned_results} for both time expression extraction and event extraction using dense and MoE models, along with compositional extraction experiments that jointly evaluate temporal and event understanding. The reported scores provide a fine-grained view of model behavior under domain shift, adversarial perturbations, vocabulary variation, and longer contextual inputs. We also analyze the effect of model scale to understand how performance changes with increasing parameter size across different evaluation settings. In addition, we report results for fine-tuned models to compare task-specific adaptation against prompting-based reasoning strategies. These detailed tables complement the main analysis by exposing performance consistency, robustness trends, and setting-specific weaknesses that are less visible from aggregated summaries alone.

\begin{table*}[t!]
\centering
\footnotesize
\setlength{\tabcolsep}{5pt}

\begin{tabular}{llcccccc}
\toprule
\textbf{Model} & \textbf{Reasoning} & \textbf{Base} & \textbf{\textit{T2C}} & \textbf{\textit{Adv}} & \textbf{Length} & \textbf{\textit{Adv\_P}} & \textbf{\textit{Voc}} \\
\midrule

\multirow{7}{*}{Llama 17B}
& D     & 0.75 & 0.83 & 0.78 & 0.75 & 0.75 & 0.73 \\
& I     & 0.82 & 0.88 & 0.89 & 0.79 & 0.77 & 0.78 \\
& A     & 0.83 & 0.86 & 0.84 & 0.79 & 0.81 & 0.84 \\
& D+I   & 0.82 & 0.82 & 0.80 & 0.69 & 0.78 & 0.79 \\
& D+A   & 0.81 & 0.82 & 0.78 & 0.79 & 0.83 & 0.84 \\
& I+A   & 0.84 & 0.87 & 0.89 & 0.79 & 0.85 & 0.89 \\
& D+I+A & 0.87 & 0.80 & 0.82 & 0.75 & 0.83 & 0.88 \\

\midrule

\multirow{7}{*}{Mistral 13B}
& D     & 0.61 & 0.76 & 0.61 & 0.49 & 0.64 & 0.63 \\
& I     & 0.70 & 0.76 & 0.67 & 0.59 & 0.74 & 0.69 \\
& A     & 0.61 & 0.78 & 0.57 & 0.23 & 0.68 & 0.59 \\
& D+I   & 0.70 & 0.75 & 0.70 & 0.63 & 0.70 & 0.74 \\
& D+A   & 0.59 & 0.73 & 0.60 & 0.19 & 0.70 & 0.58 \\
& I+A   & 0.78 & 0.79 & 0.77 & 0.48 & 0.77 & 0.77 \\
& D+I+A & 0.74 & 0.78 & 0.74 & 0.46 & 0.75 & 0.70 \\

\midrule

\multirow{7}{*}{Qwen 22B}
& D     & 0.73 & 0.84 & 0.79 & 0.77 & 0.79 & 0.80 \\
& I     & 0.80 & 0.84 & 0.88 & 0.80 & 0.83 & 0.88 \\
& A     & 0.87 & 0.86 & 0.87 & 0.76 & 0.86 & 0.87 \\
& D+I   & 0.89 & 0.75 & 0.77 & 0.78 & 0.85 & 0.87 \\
& D+A   & 0.87 & 0.86 & 0.87 & 0.77 & 0.88 & 0.87 \\
& I+A   & 0.89 & 0.87 & 0.90 & 0.80 & 0.89 & 0.90 \\
& D+I+A & 0.89 & 0.86 & 0.89 & 0.78 & 0.87 & 0.89 \\

\bottomrule
\end{tabular}

\caption{Time expression extraction results for MoE models under different reasoning strategies. Scores are reported using micro F1.}
\label{tab:moe_results}
\vspace{-8pt}

\end{table*}

\begin{table*}[t!]
\centering
\footnotesize
\setlength{\tabcolsep}{5pt}

\begin{tabular}{llcccccc}
\toprule
\textbf{Model} & \textbf{Reasoning} & \textbf{Base} & \textbf{\textit{T2C}} & \textbf{\textit{Adv}} & \textbf{Length} & \textbf{\textit{Adv\_P}} & \textbf{\textit{Voc}} \\
\midrule

\multirow{7}{*}{Llama 17B}
& D     & 0.77 & 0.24 & 0.77 & 0.66 & 0.50 & 0.71 \\
& I     & 0.82 & 0.23 & 0.82 & 0.66 & 0.51 & 0.70 \\
& A     & 0.72 & 0.22 & 0.75 & 0.54 & 0.53 & 0.74 \\
& D+I   & 0.80 & 0.24 & 0.77 & 0.60 & 0.50 & 0.70 \\
& D+A   & 0.77 & 0.23 & 0.79 & 0.43 & 0.54 & 0.77 \\
& I+A   & 0.82 & 0.26 & 0.83 & 0.60 & 0.56 & 0.81 \\
& D+I+A & 0.82 & 0.25 & 0.83 & 0.59 & 0.56 & 0.83 \\

\midrule

\multirow{7}{*}{Mistral 13B}
& D     & 0.46 & 0.45 & 0.47 & 0.21 & 0.35 & 0.48 \\
& I     & 0.64 & 0.26 & 0.63 & 0.38 & 0.49 & 0.65 \\
& A     & 0.46 & 0.28 & 0.47 & 0.07 & 0.44 & 0.45 \\
& D+I   & 0.64 & 0.27 & 0.64 & 0.38 & 0.49 & 0.64 \\
& D+A   & 0.51 & 0.25 & 0.50 & 0.09 & 0.47 & 0.50 \\
& I+A   & 0.63 & 0.21 & 0.62 & 0.19 & 0.51 & 0.62 \\
& D+I+A & 0.60 & 0.19 & 0.62 & 0.16 & 0.50 & 0.61 \\

\midrule

\multirow{7}{*}{Qwen 22B}
& D     & 0.54 & 0.43 & 0.54 & 0.36 & 0.40 & 0.58 \\
& I     & 0.71 & 0.27 & 0.70 & 0.65 & 0.54 & 0.80 \\
& A     & 0.50 & 0.45 & 0.50 & 0.30 & 0.35 & 0.51 \\
& D+I   & 0.70 & 0.28 & 0.75 & 0.52 & 0.51 & 0.68 \\
& D+A   & 0.64 & 0.37 & 0.65 & 0.15 & 0.46 & 0.64 \\
& I+A   & 0.80 & 0.28 & 0.80 & 0.51 & 0.55 & 0.80 \\
& D+I+A & 0.79 & 0.28 & 0.79 & 0.53 & 0.54 & 0.80 \\

\bottomrule
\end{tabular}

\caption{Event expression extraction results for MoE models under different reasoning strategies. Scores are reported using micro F1.}
\label{tab:moe_event}
\vspace{-8pt}

\end{table*}

\begin{table*}[t!]
\centering
\renewcommand{\arraystretch}{0.9}
\footnotesize
\setlength{\tabcolsep}{5pt}

\begin{tabular}{llcccccc}
\toprule
\textbf{Model} & \textbf{Reasoning} & \textbf{Base} & \textbf{\textit{T2C}} & \textbf{\textit{Adv}} & \textbf{Length} & \textbf{\textit{Adv\_P}} & \textbf{\textit{Voc}} \\
\midrule

\multirow{7}{*}{Llama 8B}
& D     & 0.67 & 0.79 & 0.66 & 0.59 & 0.70 & 0.67 \\
& I     & 0.76 & 0.82 & 0.75 & 0.62 & 0.76 & 0.76 \\
& A     & 0.72 & 0.83 & 0.71 & 0.30 & 0.76 & 0.72 \\
& D+I   & 0.76 & 0.82 & 0.76 & 0.62 & 0.77 & 0.78 \\
& D+A   & 0.71 & 0.83 & 0.70 & 0.30 & 0.77 & 0.71 \\
& I+A   & 0.76 & 0.83 & 0.76 & 0.58 & 0.82 & 0.77 \\
& D+I+A & 0.77 & 0.84 & 0.78 & 0.56 & 0.83 & 0.79 \\

\midrule

\multirow{7}{*}{Mistral 8B}
& D     & 0.57 & 0.74 & 0.56 & 0.64 & 0.65 & 0.56 \\
& I     & 0.85 & 0.80 & 0.83 & 0.74 & 0.77 & 0.84 \\
& A     & 0.77 & 0.82 & 0.75 & 0.68 & 0.77 & 0.75 \\
& D+I   & 0.85 & 0.80 & 0.81 & 0.73 & 0.82 & 0.84 \\
& D+A   & 0.77 & 0.81 & 0.73 & 0.75 & 0.78 & 0.76 \\
& I+A   & 0.84 & 0.84 & 0.86 & 0.78 & 0.78 & 0.82 \\
& D+I+A & 0.86 & 0.84 & 0.86 & 0.76 & 0.82 & 0.85 \\

\midrule

\multirow{7}{*}{Qwen 7B}
& D     & 0.76 & 0.81 & 0.75 & 0.58 & 0.75 & 0.76 \\
& I     & 0.83 & 0.85 & 0.81 & 0.70 & 0.81 & 0.84 \\
& A     & 0.73 & 0.85 & 0.74 & 0.22 & 0.68 & 0.54 \\
& D+I   & 0.84 & 0.85 & 0.81 & 0.71 & 0.81 & 0.85 \\
& D+A   & 0.73 & 0.85 & 0.73 & 0.21 & 0.65 & 0.60 \\
& I+A   & 0.80 & 0.83 & 0.79 & 0.44 & 0.67 & 0.55 \\
& D+I+A & 0.79 & 0.83 & 0.78 & 0.37 & 0.69 & 0.58 \\

\bottomrule
\end{tabular}

\caption{Time expression extraction results for dense models under different reasoning strategies. Scores are reported using micro F1.}
\label{tab:dense_temporal}
\vspace{-8pt}

\end{table*}

\begin{table*}[t!]
\centering
\footnotesize
\setlength{\tabcolsep}{5pt}

\begin{tabular}{llcccccc}
\toprule
\textbf{Model} & \textbf{Reasoning} & \textbf{Base} & \textbf{\textit{T2C}} & \textbf{\textit{Adv}} & \textbf{Length} & \textbf{\textit{Adv\_P}} & \textbf{\textit{Voc}} \\
\midrule

\multirow{7}{*}{Llama 8B}
& D     & 0.55 & 0.35 & 0.52 & 0.35 & 0.46 & 0.54 \\
& I     & 0.74 & 0.27 & 0.75 & 0.48 & 0.54 & 0.76 \\
& A     & 0.54 & 0.29 & 0.57 & 0.33 & 0.49 & 0.55 \\
& D+I   & 0.75 & 0.28 & 0.75 & 0.46 & 0.55 & 0.75 \\
& D+A   & 0.51 & 0.21 & 0.48 & 0.08 & 0.34 & 0.50 \\
& I+A   & 0.63 & 0.26 & 0.66 & 0.33 & 0.50 & 0.65 \\
& D+I+A & 0.66 & 0.26 & 0.68 & 0.26 & 0.52 & 0.64 \\

\midrule

\multirow{7}{*}{Mistral 8B}
& D     & 0.65 & 0.41 & 0.66 & 0.49 & 0.50 & 0.64 \\
& I     & 0.76 & 0.38 & 0.76 & 0.57 & 0.52 & 0.78 \\
& A     & 0.48 & 0.36 & 0.52 & 0.38 & 0.41 & 0.48 \\
& D+I   & 0.76 & 0.34 & 0.76 & 0.54 & 0.53 & 0.76 \\
& D+A   & 0.65 & 0.30 & 0.64 & 0.05 & 0.48 & 0.65 \\
& I+A   & 0.75 & 0.30 & 0.75 & 0.50 & 0.52 & 0.75 \\
& D+I+A & 0.68 & 0.31 & 0.75 & 0.43 & 0.44 & 0.68 \\

\midrule

\multirow{7}{*}{Qwen 7B}
& D     & 0.53 & 0.27 & 0.54 & 0.23 & 0.48 & 0.55 \\
& I     & 0.68 & 0.29 & 0.70 & 0.29 & 0.48 & 0.69 \\
& A     & 0.35 & 0.24 & 0.37 & 0.06 & 0.30 & 0.26 \\
& D+I   & 0.65 & 0.25 & 0.66 & 0.24 & 0.48 & 0.65 \\
& D+A   & 0.49 & 0.22 & 0.51 & 0.05 & 0.29 & 0.36 \\
& I+A   & 0.56 & 0.20 & 0.60 & 0.07 & 0.43 & 0.42 \\
& D+I+A & 0.56 & 0.19 & 0.61 & 0.07 & 0.41 & 0.41 \\

\bottomrule
\end{tabular}

\caption{Event expression extraction results for dense models under different reasoning strategies. Scores are reported using micro F1.}
\label{tab:dense_event}
\vspace{-8pt}

\end{table*}

\begin{table*}[t!]
\centering
\renewcommand{\arraystretch}{0.88}
\footnotesize
\setlength{\tabcolsep}{6pt}

\begin{tabular}{llcccc}
\toprule
\textbf{Model} & \textbf{Reasoning} & \textbf{Base} & \textbf{\textit{Adv}} & \textbf{\textit{T2C}} & \textbf{Length} \\
\midrule

\multirow{7}{*}{Mistral 3B}
& D     & 0.60 & 0.59 & 0.44 & 0.36 \\
& I     & 0.68 & 0.64 & 0.44 & 0.46 \\
& A     & 0.28 & 0.28 & 0.41 & 0.13 \\
& D+I   & 0.67 & 0.65 & 0.35 & 0.41 \\
& D+A   & 0.31 & 0.29 & 0.30 & 0.05 \\
& I+A   & 0.39 & 0.37 & 0.32 & 0.22 \\
& D+I+A & 0.57 & 0.60 & 0.35 & 0.21 \\

\midrule

\multirow{7}{*}{Mistral 24B}
& D     & 0.71 & 0.71 & 0.32 & 0.57 \\
& I     & 0.80 & 0.80 & 0.33 & 0.61 \\
& A     & 0.65 & 0.64 & 0.30 & 0.45 \\
& D+I   & 0.79 & 0.78 & 0.30 & 0.54 \\
& D+A   & 0.69 & 0.69 & 0.29 & 0.22 \\
& I+A   & 0.79 & 0.79 & 0.27 & 0.55 \\
& D+I+A & 0.79 & 0.79 & 0.26 & 0.56 \\

\midrule

\multirow{7}{*}{Mistral 123B}
& D     & 0.79 & 0.80 & 0.32 & 0.69 \\
& I     & 0.82 & 0.82 & 0.32 & 0.71 \\
& A     & 0.56 & 0.59 & 0.42 & 0.43 \\
& D+I   & 0.82 & 0.82 & 0.31 & 0.70 \\
& D+A   & 0.76 & 0.78 & 0.29 & 0.56 \\
& I+A   & 0.82 & 0.83 & 0.31 & 0.70 \\
& D+I+A & 0.82 & 0.83 & 0.31 & 0.69 \\

\bottomrule
\end{tabular}

\caption{Event expression extraction results for model size analysis. Scores are reported using micro F1.}
\label{tab:size_event}
\vspace{-8pt}

\end{table*}

\begin{table*}[t!]
\centering
\renewcommand{\arraystretch}{0.88}
\footnotesize
\setlength{\tabcolsep}{6pt}

\begin{tabular}{llcccc}
\toprule
\textbf{Model} & \textbf{Reasoning} & \textbf{Base} & \textbf{\textit{Adv}} & \textbf{\textit{T2C}} & \textbf{Length} \\
\midrule

\multirow{7}{*}{Mistral 3B}
& D     & 0.62 & 0.60 & 0.77 & 0.52 \\
& I     & 0.78 & 0.77 & 0.79 & 0.68 \\
& A     & 0.69 & 0.68 & 0.80 & 0.49 \\
& D+I   & 0.77 & 0.77 & 0.78 & 0.67 \\
& D+A   & 0.68 & 0.64 & 0.79 & 0.50 \\
& I+A   & 0.73 & 0.73 & 0.82 & 0.63 \\
& D+I+A & 0.72 & 0.71 & 0.81 & 0.65 \\

\midrule

\multirow{7}{*}{Mistral 24B}
& D     & 0.80 & 0.79 & 0.84 & 0.70 \\
& I     & 0.88 & 0.87 & 0.86 & 0.76 \\
& A     & 0.85 & 0.85 & 0.86 & 0.77 \\
& D+I   & 0.88 & 0.88 & 0.87 & 0.78 \\
& D+A   & 0.84 & 0.83 & 0.87 & 0.77 \\
& I+A   & 0.89 & 0.89 & 0.88 & 0.79 \\
& D+I+A & 0.90 & 0.89 & 0.88 & 0.79 \\

\midrule

\multirow{7}{*}{Mistral 123B}
& D     & 0.85 & 0.82 & 0.86 & 0.70 \\
& I     & 0.89 & 0.86 & 0.84 & 0.76 \\
& A     & 0.84 & 0.84 & 0.86 & 0.77 \\
& D+I   & 0.88 & 0.88 & 0.83 & 0.78 \\
& D+A   & 0.82 & 0.82 & 0.87 & 0.77 \\
& I+A   & 0.89 & 0.89 & 0.87 & 0.79 \\
& D+I+A & 0.89 & 0.89 & 0.87 & 0.79 \\

\bottomrule
\end{tabular}

\caption{Time expression extraction results for model size analysis. Scores are reported using micro F1.}
\label{tab:size_temporal}
\vspace{-8pt}

\end{table*}

\begin{table*}[t]
\centering
\small
\setlength{\tabcolsep}{10pt}
\renewcommand{\arraystretch}{0.92}

\begin{minipage}[t]{0.48\textwidth}
\centering

\begin{tabular}{llc}
\toprule
\textbf{Model} & \textbf{Reasoning} & \textbf{F1} \\
\midrule

\multirow{7}{*}{Llama 17B}
& D     & 0.71 \\
& I     & 0.75 \\
& A     & 0.69 \\
& D+I   & 0.75 \\
& D+A   & 0.75 \\
& I+A   & 0.82 \\
& D+I+A & 0.80 \\

\midrule

\multirow{7}{*}{Mistral 13B}
& D     & 0.45 \\
& I     & 0.62 \\
& A     & 0.45 \\
& D+I   & 0.62 \\
& D+A   & 0.47 \\
& I+A   & 0.55 \\
& D+I+A & 0.54 \\

\midrule

\multirow{7}{*}{Qwen 22B}
& D     & 0.56 \\
& I     & 0.82 \\
& A     & 0.61 \\
& D+I   & 0.83 \\
& D+A   & 0.66 \\
& I+A   & 0.62 \\
& D+I+A & 0.83 \\

\bottomrule
\end{tabular}

\captionof{table}{Compositional extraction results for MoE models (time + event). Scores are reported using micro F1.}
\label{tab:comp_moe}

\end{minipage}
\hfill
\begin{minipage}[t]{0.48\textwidth}
\centering

\begin{tabular}{llc}
\toprule
\textbf{Model} & \textbf{Reasoning} & \textbf{F1} \\
\midrule

\multirow{7}{*}{Llama 8B}
& D     & 0.54 \\
& I     & 0.75 \\
& A     & 0.59 \\
& D+I   & 0.76 \\
& D+A   & 0.58 \\
& I+A   & 0.69 \\
& D+I+A & 0.72 \\

\midrule

\multirow{7}{*}{Mistral 8B}
& D     & 0.37 \\
& I     & 0.39 \\
& A     & 0.53 \\
& D+I   & 0.40 \\
& D+A   & 0.65 \\
& I+A   & 0.77 \\
& D+I+A & 0.77 \\

\midrule

\multirow{7}{*}{Qwen 7B}
& D     & 0.54 \\
& I     & 0.71 \\
& A     & 0.51 \\
& D+I   & 0.71 \\
& D+A   & 0.54 \\
& I+A   & 0.68 \\
& D+I+A & 0.69 \\

\bottomrule
\end{tabular}

\captionof{table}{Compositional extraction results for dense models (time + event). Scores are reported using micro F1.}
\label{tab:comp_dense}

\end{minipage}

\vspace{-8pt}
\end{table*}




\begin{table*}[t!]
\centering
\small
\setlength{\tabcolsep}{6pt}
\begin{tabular}{lcccccc}
\toprule
\multicolumn{7}{c}{\textbf{Time Expression Extraction (Fine-tuned Models)}} \\
\midrule
\textbf{Model} & \textbf{Base} & \textbf{\textit{T2C}} & \textbf{\textit{Adv}} & \textbf{Length} & \textbf{\textit{Voc}} & \textbf{\textit{Adv\_P}} \\
\midrule
RoBERTa & 0.93 & 0.69 & 0.92 & 0.88 & 0.93 & 0.79 \\
T5      & 0.90 & 0.55 & 0.89 & 0.83 & 0.90 & 0.79 \\
GPT-2   & 0.88 & 0.48 & 0.86 & 0.83 & 0.88 & 0.74 \\
\midrule
\multicolumn{7}{c}{\textbf{Event Expression Extraction (Fine-tuned Models)}} \\
\midrule
\textbf{Model} & \textbf{Base} & \textbf{\textit{T2C}} & \textbf{\textit{Adv}} & \textbf{Length} & \textbf{\textit{Voc}} & \textbf{\textit{Adv\_P}} \\
\midrule
RoBERTa & 0.85 & 0.19 & 0.82 & 0.82 & 0.85 & 0.50 \\
T5      & 0.70 & 0.17 & 0.68 & 0.66 & 0.70 & 0.39 \\
GPT-2   & 0.84 & 0.18 & 0.78 & 0.85 & 0.84 & 0.50 \\
\bottomrule
\end{tabular}
\caption{Performance of fine-tuned models on time and event expression extraction across multiple generalization dimensions. Scores are reported using micro F1.}
\label{tab:finetuned_results}
\vspace{-8pt}
\end{table*}

\end{document}